\documentclass{article}
\usepackage{frExamplee}

\usepackage[utf8]{inputenc}
\usepackage{hyperref}
\usepackage{amsmath}
\usepackage{amssymb}
\usepackage{graphicx}
\usepackage{dsfont}
\usepackage{todonotes}
\usepackage[T1]{fontenc}
\usepackage{textcomp}
\usepackage{subfigure}
\usepackage{algpseudocode}
\usepackage{algorithm}
\usepackage{makecell}
\usepackage{natbib}
\let\bibhang\relax
\usepackage{apalike}
\usepackage{setspace}
\renewcommand{\cite}{\citep}

\makeatletter
\def\thanks#1{\protected@xdef\@thanks{\@thanks
        \protect\footnotetext{#1}}}
\makeatother

\title{Air-Ground Collaboration with SPOMP: Semantic Panoramic Online Mapping and Planning}
\author{Ian D. Miller$^{1*}$, Fernando Cladera$^1$, Trey Smith$^2$, Camillo Jose Taylor$^1$, Vijay Kumar$^1$%
\thanks{$^*$ Ian D. Miller was with the University of Pennsylvania when this work was completed, and is now with Burro.  $^1$  GRASP Laboratory, University of Pennsylvania,
Philadelphia PA 19104. $^2$ NASA Ames Intelligent Robotics Group, Moffett Field, CA 94035. Corresponding author: Ian D. Miller \texttt{iandm@alumni.upenn.edu}}
}

\begin{document}

\maketitle

\begin{abstract}
Mapping and navigation have gone hand-in-hand since long before robots existed.
Maps are a key form of communication, allowing someone who has never been somewhere to nonetheless navigate that area successfully.
In the context of multi-robot systems, the maps and information that flow between robots are necessary for effective collaboration, whether those robots are operating concurrently, sequentially, or completely asynchronously.
In this paper, we argue that maps must go beyond encoding purely geometric or visual information to enable increasingly complex autonomy, particularly between robots.
We propose a framework for multi-robot autonomy, focusing in particular on air and ground robots operating in outdoor 2.5D environments.
We show that semantic maps can enable the specification, planning, and execution of complex collaborative missions, including localization in GPS-denied settings.
A distinguishing characteristic of this work is that we strongly emphasize field experiments and testing, and by doing so demonstrate that these ideas can work at scale in the real world.
We also perform extensive simulation experiments to validate our ideas at even larger scales.
We believe these experiments and the experimental results constitute a significant step forward toward advancing the state-of-the-art of large-scale, collaborative multi-robot systems operating with real communication, navigation, and perception constraints.
\end{abstract}

\section{Introduction}
Multi-robot heterogeneous systems provide benefits in a wide range of tasks \cite{rizk2019cooperative}.
Consider one such task: exploratory mapping, where a robot team is tasked with mapping objects of interest within some user-defined region.
Such a use case arises in domains ranging from agriculture \cite{ribeiro2021multi, chen2020sloam} to search-and-rescue \cite{lindqvist2022multimodality}, to radiation mapping \cite{gabrlik2021automated}.
Employing robots with diverse sensing and mobility characteristics often leads to a team that is more capable, flexible, and efficient.
However, to efficiently utilize these robots, we must find a suitable representation for data communicated between robots.

It is first worth distinguishing between representations \textit{internal} to a robot and those \textit{shared} between robots.
A robot may have some set of environment representations that it uses internally, but we do not require that robots exchange that information verbatim.
Robots only share some subset of their state, not the entirety.
We call the representation used by the robot that built it \textit{internal}, and the information communicated to other robots the \textit{shared} representation.

This leads us to the following list of properties for our shared representation:
\begin{itemize}
    \item \textbf{Sufficiency:} Given a suitable task-performance function, we want to minimize the penalty for communicating the \textit{shared} representation as opposed to the full \textit{internal} one.
    \item \textbf{Efficiency:} We want to minimize the resources needed for communication, such as time, power, and bandwidth.
    \item \textbf{Commonality:} The shared representation should be usable by all robots, even those with diverse sensing and mobility.  That is, it should be invariant to variations in each robot platform.
    \item \textbf{Interpretability:} The representation should be understandable not only by robots, but also by human operators.
\end{itemize}

To help build intuition, and in light of our desired property of interpretability, it is helpful to consider how humans approach this problem.
We typically communicate not in terms of geometry, but rather in terms of objects and their meaning for a particular task.
For instance, we might communicate a room layout by describing what wall or corner a piece of furniture should be against, not in terms of cartesian coordinates or a full 3D model of the space.
Because this information is encoded in terms of human language, we refer to it as \textit{semantics}.

We argue that human language is not simply convenient, but actually reflects more fundamental properties of the environment.
Semantics naturally represent properties that are invariant under a wide variety of transformations.
A chair remains a chair under translation, rotation, scaling, color shifting, sensing, \textit{et cetera}.
Its ``chair-ness'' is an intrinsic property not dependent on any external factor.
This invariance is captured by our properties of commonality and interpretability, and even sufficiency and efficiency since encoding only invariant properties throws out extraneous information that may not be useful to other agents and expensive to share.

Thus far, we have taken a very general definition of semantics.
However, in practice, most works have encoded semantics into an environment by labeling geometry or objects with some set of semantic labels, or possibly a probability distribution over some set \cite{garg2020semantics}.
The choice of this set of labels is highly dependent on the tasks a robot is intended to complete, and has the benefit of being human-interpretable by definition.
For instance, classes such as \texttt{graspable} or \texttt{slippery} might be relevant to a robotic manipulator, classes such as \texttt{coffee mug} or \texttt{towel} relevant to a home assistant robot, and classes such as \texttt{road} or \texttt{pedestrian} relevant to an autonomous vehicle.
In recent years, deep learning methods for rapidly and accurately labeling images and even point clouds based on training data have become widely adopted.
We therefore leverage these methods throughout this work and adopt a set of classes for this work that are distinctive, detectable from the air and ground through cameras or LiDAR, and relevant for traversability, thereby ensuring that we meet all the requirements for our representation.

In this paper, we present an integrated system for air-ground robotic collaboration built on top of semantic maps.
Like our prior work \cite{miller2022stronger}, we adopt the architecture of an aerial robot with multiple ground robots consuming the aerial data.
We select this architecture because it allows us to leverage the unique capabilities of aerial and ground robots.
Aerial robots can fly at high altitude to view large portions of the environment, as well as explore more rapidly and with less concern for obstacles.
In addition, due to their altitude, aerial vehicles can often have line-of-sight to multiple ground robots, thereby naturally being an effective communication relay.
Because of these properties, our architecture is designed to primarily exploit an aerial robot as an initial explorer and communication relay, and ground robots for more close-up investigation.

A distinguishing characteristic of our work is our emphasis on semantics at every level of autonomy.
We adopt a cross-view localization system built on semantics from our prior work \cite{miller2021any}, but also employ semantics for ground robot global planning.
Secondly, we heavily employ integrated depth panoramas constructed by our LiDAR odometry algorithm, LLOL \cite{qu2022llol}.
These panoramas are semantically segmented and used for localization, and are also used directly for local terrain assessment and local planning.
By building our approach around semantics, depth panoramas, and aerial maps, we simplify our overall autonomy stack around these core representations, all of which are compact and efficient to use.

We devote a large portion of this paper to the experimental verification and validation of our approach.
While our primary contribution of this work is an autonomy framework which can be the foundation for a variety of tasks, we here focus on a particular scenario to experimentally study and characterize the system.
In this scenario, the overall objective of the team is to visit all clusters of cars in a predefined region with at least one ground robot.
We adopt this particular mission because it is motivated by real-world search and rescue applications and it leverages ground and aerial robots.
It is also straightforward to compute informative performance metrics for this task, such as the number of clusters successfully visited and the time taken to do so.
However, we note that this scenario is not our goal \textit{in itself}, rather, we seek to construct a more general autonomy framework with it as a motivating example.

Our contributions are as follows:
\begin{itemize}
    \item An integrated system for multi-robot collaboration, the code of which we open source (\url{https://github.com/KumarRobotics/spomp-system}).  A video demonstrating SPOMP in operation is available there as well.
    \item An efficient depth panorama-based local planner for ground robots, along with a mechanism for learning and sharing compact traversability information between ground robots while incorporating aerial context.
    \item Experiments in simulation and the real world demonstrating active autonomous collaboration between aerial and ground robots based on a single high-level mission specification.
\end{itemize}

\section{Background}

\subsection{Ground Robot Terrain Analysis and Planning}
Motion planning approaches can be broadly categorized into classical approaches and learning-based approaches \cite{dong2021review}.
Most classical approaches begin with some form of geometric map representation.
This representation can take the form of a digital elevation map (DEM), voxel grid, or pointcloud, and approaches can be purely geometric or appearance-based \cite{borges2022survey}.

Early works typically used a LiDAR for sensing and constructed an elevation map from the sensor pointcloud.
They then estimated traversability using a metric computed on that terrain \cite{howard2000real, larson2011off}.
Some more recent methods, recognizing the inherent polar nature of 3D LiDAR, instead directly compute traversability on the points of an organized sweep \cite{reddy2016computing}.
This approach takes advantage of the adjacency of points in an organized point cloud.
However, approaches that operate on these point-based representations ultimately bin these points into a grid, either cartesian \cite{neuhaus2009terrain} or polar \cite{martinez2020reactive}, for planning purposes.

While geometric algorithms are straightforward and work well in many environments, not all obstacles are characterized by geometry alone.
For instance, tall grass may look like an obstacle but be easily driven over, and muddy or sandy regions may appear smooth but be treacherous nonetheless.
3D LiDARs are also expensive and complex compared to cameras.
To address these challenges, many have proposed vision-based classification systems \cite{sevastopoulos2022survey}.
One early use of vision leveraged LiDAR to compute a traversable region, and then used vision to extrapolate that region to parts of an image with a similar appearance \cite{thrun2006stanley}.
More recent approaches use supervised learning in conjunction with convolutional neural networks to directly classify images \cite{hosseinpoor2021traversability}.
Some authors have recognized the complementary benefits of geometric LiDAR and learning-based image classification and proposed methods for fusing the two in a hybrid system \cite{zhou2022terrain} or directly learning traversability from LiDAR \cite{shaban2022semantic} on the basis of human labels.
\citet{schilling2017geometric} follow the idea of fusing LiDAR and vision, but also employ a semantic segmentation network as a pre-processing step for the image data.
Semantic class, therefore, can be a useful input to the traversability model, an idea which we employ in this work.

Supervised learning-based methods typically rely on human labels of traversability.
This human labelling is time consuming as well as non-transferable between different robots which may have very different mobility constraints.
As a result, many works have sought to develop methods for self-supervision of traversability prediction models, where the labels are provided by a robot's own experience.
In an early work \cite{stavens2012self}, a ground robot statistically learns the roughness of terrains based on its interoceptive sensing.
Another method \cite{ho2013traversability} learns a kernel function to predict what a robot's configuration would be on a given region of terrain.
\citet{pragr2019aerial} learn a heatmap of traversability on top of a satellite image on the basis of experienced data.
More recently, methods have been proposed to learn traversability based on data from teleoperating the robot using a deep neural network \cite{wellhausen2019should}.
End-to-end approaches have also been proposed, such as ViKiNG \cite{shah2022viking}, which does not explicitly model traversability on a per-pixel or per-voxel basis, but rather directly determines the best plan based on imagery.
ViKiNG additionally constructs a topological graph as it explores to aid in global planning.
This approach builds on the work of other authors that have sought to build roadmaps on top of occupancy grid maps \cite{park2012efficient}.
These traversability graphs are much sparser than a dense grid map, enabling more rapid planning, and our approach takes advantage of this.

A logical next step for self-supervised learning-based methods is to not only build a training dataset from a robot's experience, but to train the robot online \textit{during} its experiences.
These methods can use Bayesian clustering \cite{lee2017incremental} or a Multi-layer Perceptron (MLP) \cite{frey2023fast} to rapidly train online on the basis of experienced data.
Both approaches perform image segmentation into superpixels \cite{lee2017incremental} or learned segments \cite{frey2023fast} as a preprocessing step.
Our approach follows this idea on online learning with an MLP, but for the purpose of extrapolating classical geometric terrain analysis to areas that only have aerial UAV imagery.

In our work, we draw inspiration from approaches that operate directly on organized LiDAR sweeps.
However, we avoid binning into grid cells and instead plan directly in this space.
In addition, we incorporate ideas from learning and traversability graphs in order to use semantically labeled aerial maps to extrapolate local terrain experience across the global map.
We perform this fitting and extrapolation process in real-time, feeding it back into our planner for constant refinement.
Unlike these works, we also incorporate information from other robots as part of the online estimation and develop a compact representation which can be efficiently shared between robots.

\subsection{Air-Ground Teaming}

It has long been recognized that heterogeneous teams of air and ground robots can achieve better performance than either robot individually.
Several authors have developed a taxonomy for different roles that Unmanned Aerial Vehicles (UAVs) and Unmanned Ground Vehicles (UGVs) can assume in a team, including acting as a sensor, actuator, or playing an auxiliary role \cite{ding2021review, liu2022review}.

Some of the most recent large-scale work on air/ground collaboration has come from the DARPA SubT Challenge \cite{chung2023into, tranzatto2022cerberus, morrell2022nebula, hudson2021heterogeneous}.
These teams all utilized heterogeneous robot teams to locate various objects in diverse underground environments in a task similar to ours.
Nonetheless, the challenges and constraints of underground operation are very different from those on the surface; the domain we focus on here.
In particular, the advantage of a quadrotor as an eye-in-the-sky is reduced in favor of its utility in visiting difficult to reach areas.
There is no way to obtain a high-altitude view of a tunnel system, but we argue that if possible for a given environment, doing so is highly valuable.
In our work, we employ the UAV as both a sensor for gathering an aerial map as well as an active communication relay.
We here discuss prior work addressing both of these roles.

\subsubsection{UAV for Mapping}
Some authors have used UAVs and UGVs in cooperative mapping in relatively homogenous ways \cite{yue2020collaborative, scherer2021resilient}.
In these works, the UAVs fly at a relatively low altitude and use similar sensing to UGVs.
Map representations, sensing, and viewpoints are all similar, making map fusion primarily between homogeneous maps.
However, the heterogeneity of mobility of both types of robots can be used to reach different parts of the environment.

By contrast, many authors have recognized the particular strengths of UAVs in high-altitude flight, gaining a perspective with a speed that a UGV cannot compete with.
One approach \cite{vandapel2006unmanned} takes LiDAR maps built from a crewed helicopter and uses them for localizing a UGV equipped with LiDAR, as well as for planning its global trajectories.
Other approaches have replaced the expensive and heavy LiDAR sensor on the aerial platform with a vision-equipped UAV.
Some of these approaches operate at relatively small scales.
In one work \cite{su2021framework}, a UAV identifies targets for a UGV to track.  However, the UGV planning and localization operates completely independently of the UAV, rather than using the UAV only to identify goals.
\citet{fankhauser2016collaborative} similarly operate at a small scale, but perform localization and mapping cooperatively on both robots.
However, the quadrotor flies low enough that image features can be directly registered between robots, and the resulting elevation map is high enough resolution for planning on it directly.

Other works perform experiments at larger scales.
A major early work \cite{grocholsky2006cooperative} employ a team of fixed-wing UAVs and UGVs to localize targets, but use GPS for localization.  Obstacles are not considered.
More recently, \citet{lazna2018cooperation} use UAV data for UGV planning, but construction of the aerial map is a computationally demanding procedure that must be run offline.
\citet{peterson2018online} build an orthomap online which they use to localize obstacles for the UGV to avoid, but communication is always assumed to be available.

Our work operates completely online and does not require constant communication.
In addition, we employ the aerial map for planning and localization, but do not specify traversability \textit{a priori}, and rather allow UGVs to learn to generalize their experience using this map.
We also focus on large-scale experiments, both in terms of area covered as well as team size.

\subsubsection{UAV for Communication}

UAVs also have properties highly conducive to facilitating communication \cite{li2021survey}.
They can operate at high altitude where line-of-sight to multiple UGVs is feasible, and can also travel more quickly without needing to worry about obstacle avoidance.
Nonetheless, algorithms are needed to determine where UAVs should go to maximize team communication and performance.

Some authors formulate this as the relay placement problem.
In a simple form, this problem can be expressed as follows: given a number of agents in the environment, select where to place relay nodes to keep the network fully connected \cite{mox2022learning}.
Other authors have generalized the problem further, such as having a number of set task points and robot points that a single UAV is tasked with visiting \cite{ding2019precedence}.
This then turns into a variant of the Travelling Salesman Problem (TSP).

However, both of these authors assume that the positions of the task and/or robot points are always known.
If communication is not constant, this is not a realistic assumption.
Some methods have sought to address this problem.
In one approach \cite{chamseddine2016communication}, the UAV uses the bearing and signal strength from different UGVs to try to optimize its position.
In another work \cite{wu2020mobility}, the UAV explicitly tries to estimate the UGV positions using a filter-based approach.
Nonetheless, both of these methods rely on some form of detection of ground robots, and if communication is lost completely may struggle to determine where to go to regain a connection.

In our work, we do not assume or try to always maintain connectivity.
Rather, we seek to balance the UAV's time between exploring new regions and acting as a relay for UGVs.
To keep track of UGV locations, the UGVs communicate their current position and current goal and the UAV uses this last-known information to try to revisit robots.
Our approach therefore makes no assumptions about the communication model, connectivity graph, or current knowledge of robot positions, and we argue that this makes it particularly robust in real-world deployments.

\subsection{Authors' Prior Work}

In this work, we employ a similar high-level design to our prior work \cite{miller2022stronger}.
This includes a modified version of the localizer from our works \cite{miller2021any, miller2022stronger}, the communication system from \cite{miller2022stronger, miller2020mine, cladera2023enabling}, the aerial mapper from \cite{miller2021any}, and the aerial planner from \cite{cladera2023enabling}.
Here, we focus on our new approaches for UGV terrain mapping, navigation, and learning from the aerial map, though we provide summaries of modules previously described in other works.
Nevertheless, our emphasis in this paper is on the \textit{integrated air/ground system}, and not purely on the terrain analysis component. 
Our experiments are much more extensive than our prior work, incorporating 3 robots across two real-world diverse and challenging environments, totalling 17.3 km autonomously travelled by the UGVs.
We additionally perform many ablation and scaling experiments in simulation to further validate our approach.

\section{Method}
\begin{figure}
    \centering
    \includegraphics[width=\textwidth]{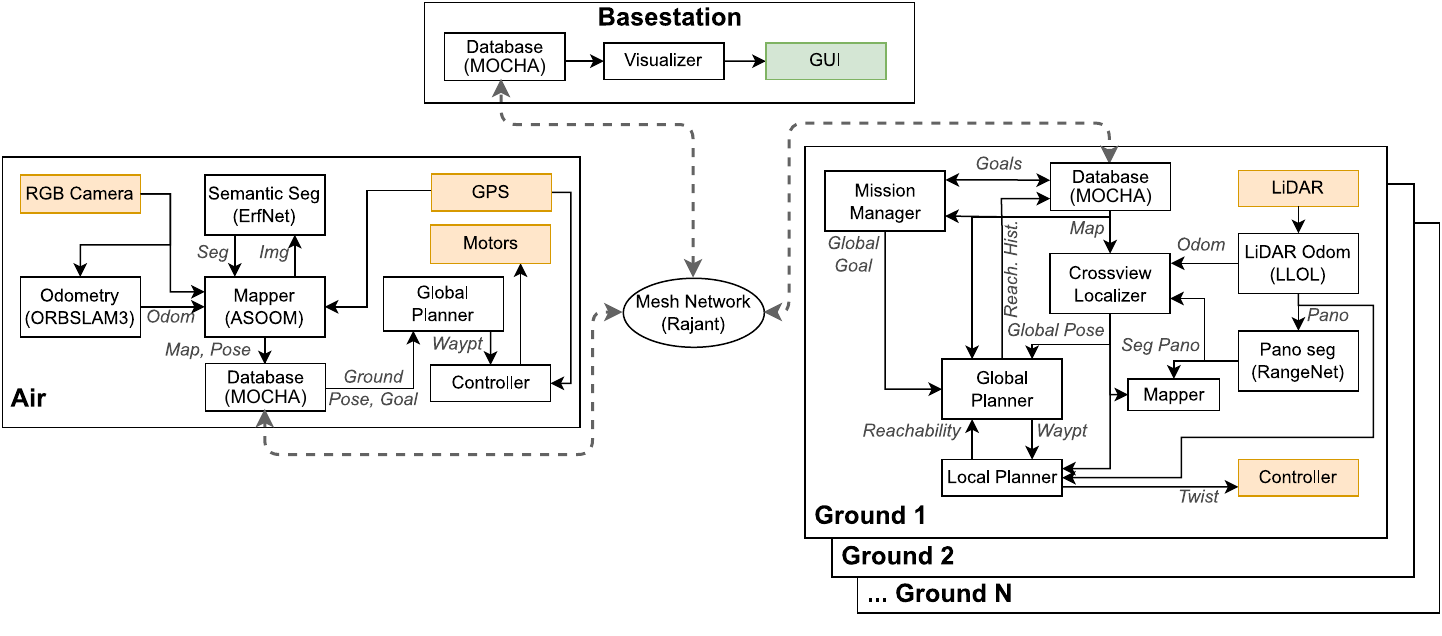}
    \caption{The SPOMP architecture described in terms of ROS nodes.  Sensors and actuators are shown in orange and visualization outputs in green.}
    \label{fig:spomp_system}
\end{figure}

A simplified overall block diagram of our system, SPOMP (Semantic Panoramic Online Mapping and Planning), is shown in Fig.~\ref{fig:spomp_system}.
The aerial robot is responsible for building a semantically segmented orthomap of the target region.
As this map is being constructed, updated maps are transmitted to ground robots when they happen to be in communication range using our communication framework and distributed database MOCHA \cite{cladera2023enabling}.
Ground robots use this semantic aerial map to localize using the cross-view localization algorithm described in our prior works \cite{miller2021any, miller2022stronger}.
Robots opportunistically share data with each other using MOCHA when they are able to.
All robots know their own poses in a common coordinate frame, and ground robots are in possession of the aerial orthomap.
However, for the system to be autonomous, we additionally need algorithms for the ground and aerial robots to make decisions about where they can and should travel, both locally and globally.
In Section \ref{sec:aerial_autonomy} we first discuss the aerial autonomy stack, beginning with the aerial mapper ASOOM in Sec.~\ref{sec:aerial_mapper} and moving to the aerial planner in Sec.~\ref{sec:aerial_planner}.
We then move to the ground autonomy stack in Section \ref{sec:ground_autonomy}, beginning by describing the high level mission manager and goal selector (Section \ref{sec:high_level_coordination}) and moving to the LiDAR odometry, segmentation (Section \ref{sec:ground_odometry_and_semantic_segmentation}), and cross-view localization algorithms (Section \ref{sec:crossview_localizer}) that provide a common reference frame to the robots.
We conclude with the UGV local and global planners responsible for planning to the selected goals while remaining safe (Sections \ref{sec:local_ground_planner} and \ref{sec:global_ground_planner}, respectively).

\subsection{Aerial Autonomy}
\label{sec:aerial_autonomy}
\subsubsection{Aerial Mapper}
\label{sec:aerial_mapper}
We employ the same Aerial Semantic Online Ortho-Mapping (ASOOM) algorithm first presented in our prior work \cite{miller2022stronger}.
ASOOM uses odometry from ORBSLAM3 \cite{campos2021orb} and GPS position in order to construct a semantic orthomap.
This is the only use of GPS in our system.
The orthomap includes several layers, including RGB color, semantic class from ErfNet \cite{romera2017erfnet}, and elevation.
Each layer is \texttt{png} compressed before being transmitted to other robots, resulting in the map size being reduced by about a factor of 10.

\subsubsection{Aerial Planner}
\label{sec:aerial_planner}
Our aerial planner is addressed in more detail in our prior work \cite{cladera2023enabling}, but we provide a brief overview here.
To exploit the UAV strengths of viewpoint and speed, we assign it two goals:
\begin{enumerate}
    \item Explore the environment to build a semantic orthomap for UGV navigation.
    \item Facilitate communication both to transmit the orthomap to UGVs as well as relay data between ground agents.
\end{enumerate}
These goals may be contradictory: if the UAV is only exploring, it will be in new regions that the UGVs have never been, and therefore never have the opportunity to send them the aerial orthomap.
By contrast, if the UAV is only facilitating communication, it will stay with the UGVs and never explore new regions.
In either scenario, the team is ineffective.

\begin{figure}
    \centering
    \includegraphics[width=0.6\textwidth]{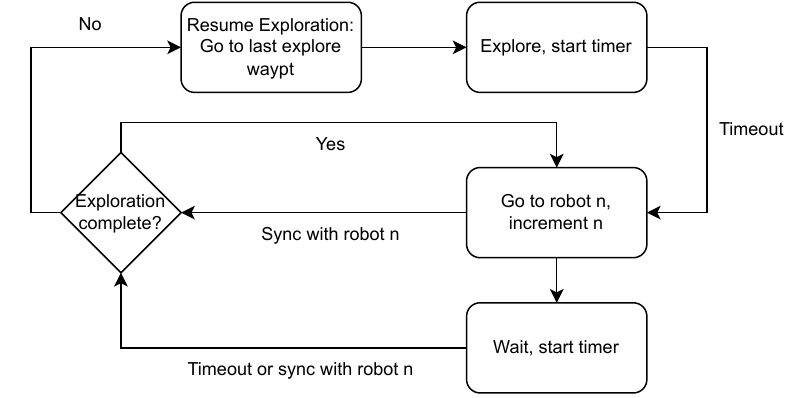}
    \caption{Simplified state machine determining the UAV high level goal.}
    \label{fig:air_router_smach}
\end{figure}

The simplified state machine for the aerial planner is shown in Fig.~\ref{fig:air_router_smach}.
We employ a time-sharing system, where the UAV alternates between tasks, assigning a fixed amount of time to each.
In exploration mode, the robot follows a preset collection of waypoints.
When the timer expires, the UAV seeks out a UGV on the basis of its knowledge of the UGV's last known location and goal.
The target UGV is selected in round-robin fashion.
If more recent information is received en route, the UAV adjusts accordingly.
Once either all data in the distributed communication system has been synchronized with the target UGV, or the UAV times out, it returns back to its last exploration waypoint, restarts the timer, and resumes exploration.
Once exploration has completed, the UAV simply spends all its time flying between the UGVs sequentially.

\subsection{Ground Autonomy}
\label{sec:ground_autonomy}
\subsubsection{High-Level Coordination}
\label{sec:high_level_coordination}
The UGVs coordinate at the level of goal locations $g \in \mathbb{R}^2$.
These goal locations can either be manually specified by an operator or selected automatically from the aerial map.
If manually specified, the goals are added to MOCHA \cite{cladera2023enabling} and then distributed by MOCHA through the mesh network to all of the robots.
If automatically detecting goals, each robot identifies clusters of the target class (vehicles in our experiments) in the aerial semantic map, and shares the goals that it has \textit{identified} via MOCHA.
Clusters are \textit{identified} by performing a morphological close followed by a morphological open operation on the binary class mask in order to remove noise and combine groups of vehicles.
We then select the goal to be the point furthest from an obstacle within a threshold distance of a cluster in order to avoid placing goals inside buildings or other unreachable areas.

Robots also share each goal's status: whether a goal is \textit{claimed}, \textit{visited}, or only \textit{identified}.
A \textit{claimed} goal is a goal that a robot is currently \textit{en route} to, but has not yet \textit{visited}.
Goals from other robots that are within a threshold distance (10 meters in our case) of a robot's goal are considered to be the same goal for the purposes of deconfliction.
The goal detection procedure is intentionally done on each robot in order to make the system entirely distributed, though goals can be provided from a central source if desired.

\begin{figure}
    \centering
    \includegraphics[width=0.8\textwidth]{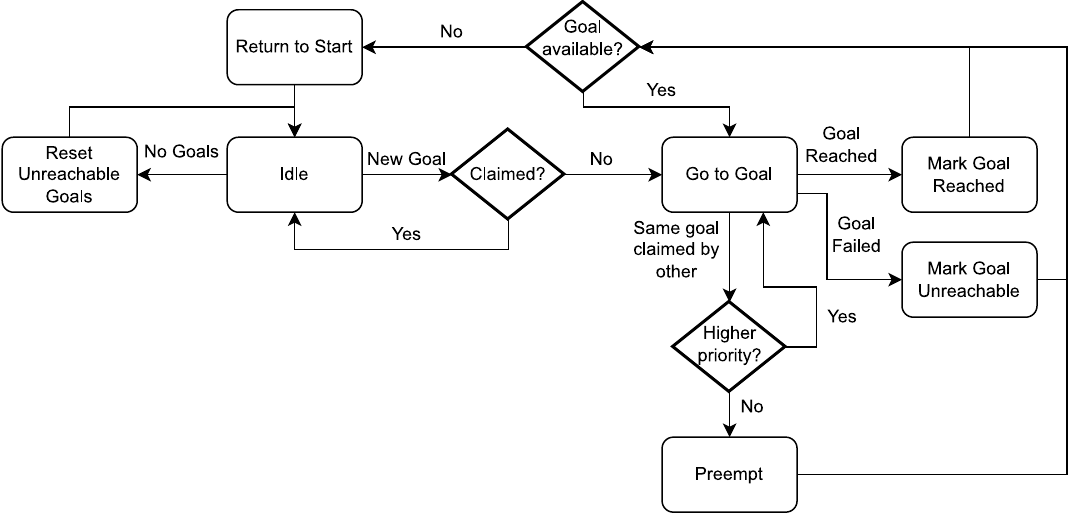}
    \caption{State machine describing the algorithm for assigning goals to ground vehicles.}
    \label{fig:coord_state_machine}
\end{figure}

We consider a mission to be completed when all goal locations have been visited by at least one UGV.
A given UGV will always select the closest goal in cartesian distance from itself that is not already claimed by another robot.
We implement a priority mechanism to disambiguate goals, with priorities pre-assigned at mission start: if a robot learns that another robot is going to the same goal, it will switch to a new goal if the other robot has a higher priority.
If a UGV is unable to find a path to a goal, or exceeds a time limit while trying to get there, it will mark the goal as unreachable and release it for other agents to attempt to reach.
This procedure is described as a state machine in Fig.~\ref{fig:coord_state_machine}.

\subsubsection{Ground Odometry and Semantic Segmentation}
\label{sec:ground_odometry_and_semantic_segmentation}
The basis for the ground odometry stack is a modified\footnote{The source code for this modified version can be found here: \url{https://github.com/versatran01/rofl-beta}} version of LLOL (Low-Latency Odometry for spinning LiDARs) \cite{qu2022llol}.
LLOL takes as input the raw LiDAR sweeps and IMU and outputs per-sweep odometry as well as integrated depth and intensity panoramas approximately every meter that the robot moves.
These panoramas can be thought of as virtual super-resolution LiDAR sweeps.

\begin{figure}
    \centering
    \includegraphics[width=0.9\textwidth]{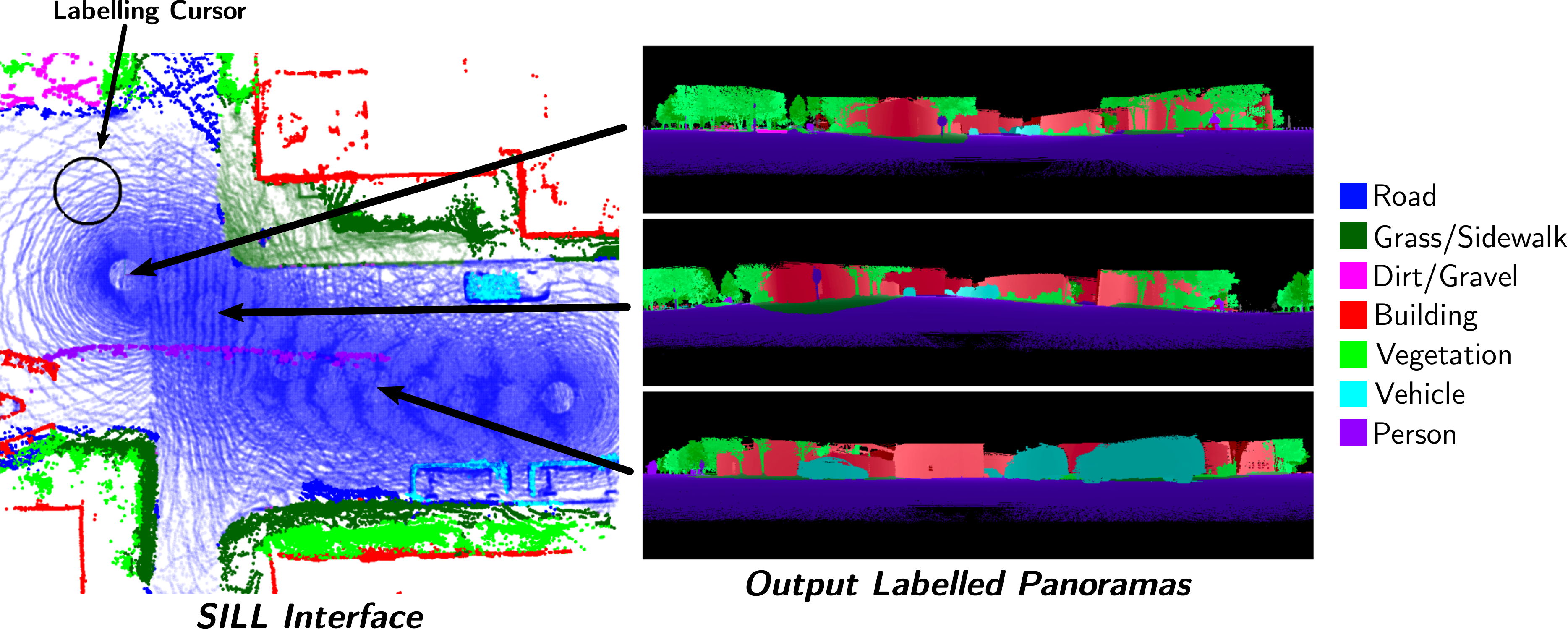}
    \caption{The Semantic Integrated LiDAR Labelling (SILL) tool and output labelled depth panoramas.}
    \label{fig:sill}
\end{figure}

We then segment these sweeps using RangeNet++ \cite{milioto2019rangenet++}, also incorporating intensity as an input.
The classes we use are shown in Fig.~\ref{fig:sill}.
The \texttt{Person} class is ignored by all downstream processing.
We make several modifications to the network.
We first shrink the network size using the same modified structure as \citet{liu2023active}.
Secondly, we replace the normalized x, y, and z coordinates with x, y, and z components of the surface normal at that point.
We also upweight the loss with range to encourage the network to more accurately classify more distant and sparse points.

In order to generate training data, we developed a custom labeling tool, SILL (Semantic Integrated LiDAR Labelling)\footnote{Source: \url{https://github.com/iandouglas96/sill}}.
The SILL interface and output labeled panoramas are shown in Fig.~\ref{fig:sill}.
SILL operates by first using LLOL to integrate panoramas into a large pointcloud.
This pointcloud is then downsampled by voxel filtering.
The key idea of SILL is to label from a top-down perspective, slicing at varying points on the $z$ axis.
Consider the example of labeling an outdoor scene.
The user first selects a $z$ value just above ground level, causing SILL to only display points at or below the ground.
The user can therefore focus on labeling just the various types of ground before increasing the $z$ threshold, at which point they can easily label above-ground objects.
This is because points which have already been labeled at lower $z$ are frozen and will not be overwritten by the new classes, making it easy to label overhanging vegetation without changing the class of the road beneath.
Using SILL, we were able to rapidly label training data from our test environments.

\subsubsection{Cross-View Localizer}
\label{sec:crossview_localizer}
The primary modification made to the localizer \cite{miller2022stronger} was to operate on integrated semantic panoramas instead of each LiDAR sweep.
This significantly reduces the computational load, because even though the higher resolution panoramas take longer to segment, they are generated far less frequently than the 10 Hz LiDAR sweeps.
Additionally, downstream processes such as localization run less often.
Because new panoramas are generated less frequently and higher frequency pose estimates are needed for planning and control, we rely on the LLOL odometry in between panoramas.

\subsubsection{Local Ground Planner}
\label{sec:local_ground_planner}
A key design decision that we made was to use depth panoramas as much as possible for planning.
While it is traditional to construct an occupancy grid or elevation map and plan on that representation \cite{papadakis2013terrain}, by directly planning in panorama space we are able to optimize the local planner by parallelizing operations on a per-row basis, increasing computation speed.
When parallelized, the entire terrain analysis pipeline takes approximately 3.5 ms per-panorama on an AMD Ryzen 3900X.
The nature of 3D spinning LiDARs inherently results in higher resolution data close to the robot and coarser data further away, which is convenient due to regions near the robot being more important for planning.
We therefore avoid the loss of data, particularly near the robot, that would result from binning points into a cartesian grid.

Let $p = (p_e, p_a)$ be a pixel coordinate on the panorama, where $p_a$ and $p_e$ are the coordinates in the azimuthal and elevation directions respectively.
Then $\rho(p) = \rho(p_e, p_a)$, $\theta(p)$, and $\phi(p)$ are the range, elevation, and azimuth at that point respectively, and $x(p)$, $y(p)$, and $z(p)$ the cartesian coordinates.
We also let $r(p) = \sqrt{x^2 (p) + y^2 (p)}$, and let the panorama have dimension $E \times A$.
Finally, we assume that there is noise in the depths of the panorama of order $\eta$.
We decompose this into $\eta_z(p) = \eta \sin(\theta(p))$ and $\eta_{xy}(p) = \eta \cos(\theta(p))$, the components perpendicular to and parallel to the $xy$ plane.

\begin{figure}
    \centering
    \includegraphics[width=\textwidth]{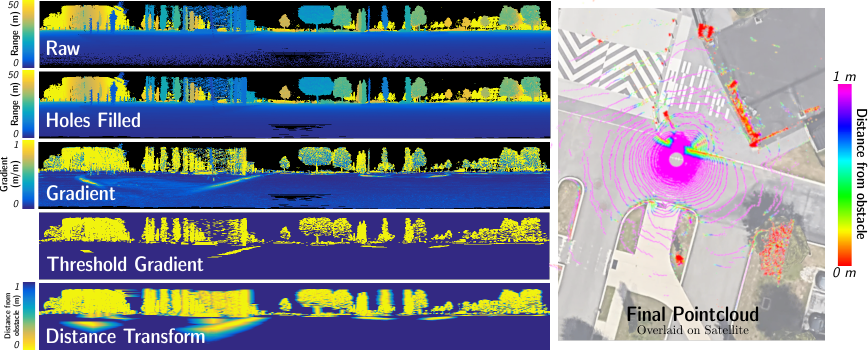}
    \caption{Visualization of the panorama after each processing step, and the final projected obstacle point cloud.}
    \label{fig:pano_obstacle}
\end{figure}

Terrain analysis consists of a number of steps operating on each panorama.  We consider each step in more detail here, and the process is illustrated in Fig.~\ref{fig:pano_obstacle}.

\paragraph{Fill holes and project:}
The depth panorama is typically of higher resolution than the raw LiDAR data, and so therefore there are often holes in the panorama with unknown depth.
We therefore loop through each row in the panorama and fill holes below a certain threshold $D_{px}$ in size with linearly interpolated depths.
This threshold is scaled with the range of the points in question such that it can be approximately specified in meters $D_m$.
More formally, at a point $p$ we have the arc length between adjacent points in the panorama
\begin{equation}
    s(p) = (2 \pi / A) * \rho(p).
\end{equation}
Then we have
\begin{equation}
    D_{px}(p) = D_m / s(p).
\end{equation}
Once this has been completed, we compute the per-pixel cartesian projection of the depths, needed for the later gradient computation.
Both operations can be done in parallel.

\paragraph{Compute and threshold gradients:}
We now compute the gradients on a per-pixel basis.
To make this operation more efficient, we take advantage of the known adjacency in the panorama structure and compute the gradient along the azimuthal and elevation directions.
This is an approximation, because there is no guarantee that the vectors from the current point to the neighboring points in the panorama are in fact perpendicular, but it is a reasonable assumption on an approximately planar surface.
To minimize noise, we want to compute the gradient using a window of approximately $W$ meters in size.
Due to the polar nature of the coordinate system, the corresponding window size in pixels varies depending on its position in the panorama.
We therefore first compute the window sizes $w_a$ and $w_e$ for the azimuthal and elevation directions
\begin{equation}
    \begin{split}
        w_a(p) &= \min\{W / s_{xy}(p), W_{max}\} / 2 + 1 \\
        w_e(p) &= \min{\left\{ \delta \in \mathbb{N} \left| h_s * \left\lvert \frac{1}{\tan{\theta(p)}} - \frac{1}{\tan{\theta(p_e - \delta, p_a)}} \right\rvert  \geq W \right. \right\}}
    \end{split}
\end{equation}
where $s_{xy}$ is the arc length on the $x-y$ plane, in other words $s_{xy}(p) = (2 \pi / A) r(p)$ and $h_s$ is the approximate height of the LiDAR above the ground when the robot is on a flat surface.
$w_a$ is computed using the known range at $p$ to estimate the arc length, and $w_e$ is computed by assuming that the robot is on a flat surface and determining the closest distance at which successive rows of the panorama are separated by at least $W$.

With these in hand, we are now ready to compute the gradients:
\begin{equation}
    \begin{split}
        \nabla_a (p) &= \frac{\max\{\lvert z(p_e, p_a-w_a(p)) - z(p_e, p_a+w_a(p)) \rvert - \eta_z(p), 0\}}{s(p) * (w_a(p) * 2 + 1)} \\
        \nabla_e (p) &= \frac{\max\{\lvert z(p) - z(p_e - w_e(p), p_a) \rvert - \eta_z(p), 0\}}{\lvert r(p) - r(p_e - w_e(p), p_a) \rvert + \eta_{xy}(p)}
    \end{split}
\end{equation}
Note that we add and subtract the components of $\eta$ in order to yield the minimum gradient that could be present if the noise is of magnitude $\eta$.
Finally, the total gradient magnitude is given by $|\nabla(p)| = \sqrt{\nabla_a^2 (p) + \nabla_e^2 (p)}$.
We threshold the panorama at a particular gradient magnitude $\nabla_{thresh}$ to form a binary image.

\paragraph{Compute per-point closest distance to obstacle:}
At this point, we have a per-pixel panorama and point cloud labeled as obstacle or not.
We then need to inflate the obstacles for planning purposes.
Since our design goal is to avoid any sort of grid projection, we now want to efficiently compute the distance for each point to the nearest obstacle point.
We do this with successive sweeps along the altitude and then the azimuth of the panorama.
The algorithm is formally described in Alg.~\ref{alg:distance_transform}.

\begin{algorithm}
\begin{algorithmic}[1]
\Require{Depth panorama $\rho(p)$, Obstacle panorama $O(p) \in \{\texttt{true}, \texttt{false}\}$}
\Ensure{Distance transformed obstacle panorama $D(p) \in [0, D_{max}]$}
\Statex
\State $D \gets D_{max} \forall p$
\State $D_a \gets D_{max} \forall p$
\For{$p_a \in \{1, \dots, A\}$}
    \State $D_{obs} \gets \infty$
    \For{$p_e \in \{1, \dots, E\}$}
        \If{$O(p_e, p_a)$}
            \State $D_a(p_e, p_a) = 0$
            \State $D_{obs} \gets \min\{ \rho(p_e, p_a), D_{obs} \}$
        \Else
            \State $D_a(p_e, p_a) \gets \min\{| \rho(p_e, p_a) - D_{obs} |, D_a(p_e, p_a)\}$
        \EndIf
    \EndFor
\EndFor
\Statex
\Function{UpdateAz}{$p_e, p_a, p_a^{last}$}
    \State $D_{obs} \gets D_a(p_e, p_a^{last}) + | p_a - p_a^{last} | * s(p_e, p_a^{last})$
    \If{$D_{obs} > D_a(p_e, p_a)$}
        \State $p_a^{last} \gets p_a$
        \State $D(p_e, p_a) \gets \min\{ D_a(p_e, p_a), D(p_e, p_a) \}$
    \Else
        \State $D(p_e, p_a) \gets \min\{ D_{obs}, D(p_e, p_a) \}$
    \EndIf
\EndFunction
\Statex
\For{$p_e \in \{1, \dots, E\}$}
    \State $p_a^{last} \gets 1$
    \For{$p_a \in \{1, \dots, A\}$}
        \State \Call{UpdateAz}{$p_e, p_a, p_a^{last}$}
    \EndFor
    \State $p_a^{last} \gets A$
    \For{$p_a \in \{A, \dots, 1\}$}
        \State \Call{UpdateAz}{$p_e, p_a, p_a^{last}$}
    \EndFor
\EndFor
\end{algorithmic}
\caption{Distance Transform Approximation}
\label{alg:distance_transform}
\end{algorithm}

We first iterate down each column, keeping track of the last time an obstacle was encountered and the distance to it.
Note that we only care about the distance to an obstacle coming from the robot's direction, so therefore we only need to make this one column pass.
We then make two passes along each row in both directions, again adding up the distances from the last closest obstacle.
Note that this addition overestimates the actual distance to the obstacle, but in practice, this algorithm yields a good enough approximation and is very fast.

As a final step, we compute what we call the \textit{reachability scan} and denote $S(\phi) \in \mathbb{R}^+ \times \{ \texttt{true}, \texttt{false} \}$.
For each $\phi$, $S$ gives us the known safe distance that the robot can travel in that direction, and a boolean value for whether there is a known obstacle at the end of that distance or not.
$S$ is computed by scanning up the specified column of the panorama until either the distance between successive scan rows exceeds a threshold (no obstacle known) or the distance to an obstacle drops below a threshold (known obstacle).
As a result, we have a known safe region, known unsafe regions, and regions that could be safe or unsafe but we are too far away to determine.

\paragraph{Controller:}
Given a global goal $\mathbf{G}$, we must first compute a local goal for the controller.
To do this, we randomly sample uniform points $\mathbf{x}$ within the known safe region specified by $S$.
We compute the cost of a choice of local goal as
\begin{equation}
    C(\mathbf{x}) = \| \mathbf{x} - \mathbf{G} \|_2 + \alpha \frac{\| \mathbf{x} \times \mathbf{g}_{old} \|_2}{\|\mathbf{g}_{old}\|_2}
\end{equation}
where $\mathbf{g}_{old}$ is the last local goal, and $\alpha$ is some constant.
We then select the new local goal $\mathbf{g} = \text{argmin}_\mathbf{x} C(\mathbf{x})$.
A goal that is in the known safe region, close to the global goal, and close to the old last goal will have the lowest cost.
The final term incentivizes the robot to be consistent in its choice of local goal instead of, for instance, switching back and forth constantly between two possible paths around an obstacle.

For our low-level controller, we take the Dynamic Window Approach (DWA) \cite{fox1997dynamic}.
We uniformly sample the control space of the robot, which for our skid-steer platform is $(v, \omega) \in \mathbb{R}^2$, where $v$ and $\omega$ are the linear and angular velocities, respectively.
We perform this sampling around the current control inputs which naturally imposes a smoothness constraint on the controller.
We then roll out that control for a time window $T_w$ into a trajectory $\tau(t)$, $t \in [0, T_w]$ and compute a cost $C$ for each trajectory $\tau$:
\begin{equation}
    \begin{split}
        C(\tau) &= C_g(\tau) + C_o(\tau) \\
        C_g(\tau) &= \| \mathbf{g} - \tau(T_w) \|_2 + \frac{\| \mathbf{g} \times \tau(T_w) \|_2}{\| \mathbf{g} \|_2} \\
        C_o(\tau) &= \frac{1}{T_w} \sum_{0 \leq t \leq T_w} -D(\Phi(\tau(t)))
    \end{split}
\end{equation}
where $\Phi$ is a function that maps from a cartesian coordinate to the indices of a nearby point in the panorama.
When sampling a point on the trajectory, we choose a point that is slightly in front of the actual robot position.
This implicitly incentivizes the robot to turn to face in the direction of the goal.
We then act on this computed control input, and replan when we receive new odometry.

\subsubsection{Global Ground Planner}
\label{sec:global_ground_planner}
The core of our global planner is a traversability graph.
Each edge $e_i$ possesses a cost $C(e_i) = -\log{P(e_i)}$, where $P(e_i)$ is the probability of a robot successfully traversing $e_i$.
Therefore, when we solve for the lowest cost path, we are finding the path with the overall highest probability of successful traversal, assuming that the probabilities for each edge are independent.
We employ a sparse traversability graph for several reasons:
\begin{itemize}
    \item Sparsity makes graph search very fast, especially for large environments.
    \item Maintaining a traversability graph allows us to keep track of traversed edges, thereby building a roadmap of known routes.
    \item Traversability, we argue, is fundamentally a property of an edge. It encodes not whether a robot can exist at a point, but rather whether a robot can get from point A to point B.
\end{itemize}
This notion of constructing a traversability graph on top of a grid map is not new \cite{park2012efficient}.
However, we combine this graph with a learned traversability model that exists on top of the aerial orthomap, especially the semantic map.
In addition, we allow a robot to update this graph with its own and other robots' experienced data.
Here we first discuss the construction of the graph, then how it is updated from the robots' own experience, and finally our method of learning to extrapolate those experiences using the aerial map.

\begin{figure}
    \centering
    \includegraphics[width=\textwidth]{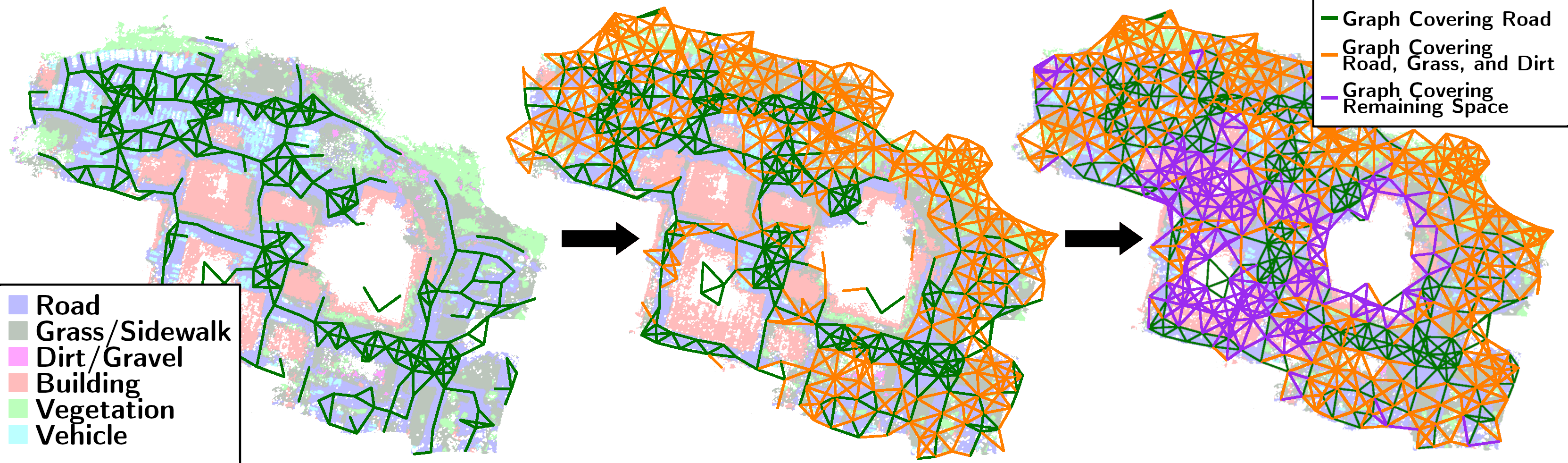}
    \caption{Visualization of the sequence of graph construction.  We begin by building a graph that well-covers the easiest to traverse class (road), and then extend the graph to additionally cover the next easiest to traverse, each class at a time.}
    \label{fig:graph_construction}
\end{figure}

\paragraph{Graph Construction:}
\label{sec:graph_construction}
We take a similar approach to graph construction as in our previous work \cite{miller2022stronger}.
For a given region that we wish to build a graph over, we sample the point that is the farthest away from the edges of the region.
We then raytrace a region around that point, which is considered covered by the graph, and repeat the sampling process ignoring points within the covered region.
This process also allows us to preserve the already-built graph as the aerial map is expanded, which is important to retain traversability data encoded in the edges.
The choice of region to cover is configurable.
We can either cover the entire known region, or we can consecutively cover regions for each class in a specified order, typically approximately from most to least traversable, as shown in Fig.~\ref{fig:graph_construction}.
The progressive method ensures that we do have edges in the regions that are traversable while keeping possible path options in case all the roads are blocked.
At the end of this process, we have both a graph and a map of the regions raytraced around each node.
We use this map to rapidly lookup accessible nodes on the graph nearby to any arbitrary point.

\paragraph{Learning Experience Extrapolation:}
It is broadly experientially true that regions that appear similar from the air are often similar on the ground.
More concretely, a region of the map that is traversable and looks like a road suggests that other road-like patches of the map are also readily traversable.
Some authors have noticed this and used this observation to fit a traversability model to local observations by a ground robot \cite{pragr2019aerial}, in this case using Locally Weighted Projection Regression (LWPR).
In a follow-up work, the authors use a haptic sensor to probe traversability, thereby learning a traversability model online \cite{pragr2023autonomous}.
Taking note of these works, and observing that we already have a large aerial map available to us, we postulate that the ground robots should be able to extrapolate their local experience to the broader map, without any \textit{a priori} information about class traversability.
We additionally observe that there is a strong predictive correlation between semantic class and traversability, as noted in Section \ref{sec:graph_construction}.
For this reason, and because we already have access to the aerial map, we employ the semantic classes as one input into our learned terrain model.
We do use an \textit{a priori} ranking of terrain classes for graph construction, but we wish to construct a more sophisticated model based on the robots' own experience.

We model the probability of traversability as $T(p) = f(\mathbf{C}(p), \mathbf{S}(p), h(p))$, where $p$ is the cell of the orthomap, $\mathbf{C}(p)$ is the RGB color vector at that point, $\mathbf{S}(p)$ is a vector of the distance to the nearest pixel of each semantic class at that point, and $h(p)$ is the elevation at that point.
Note that using the distance to all classes provides richer information to the model than the class at the point, and also allows the model to incorporate information in the local region of the point in question without needing to use convolutions.
Therefore, $f : \mathbb{R}^{3 + N_C + 1} \mapsto [0, 1]$, where $N_C$ is the number of semantic classes.
We approximate $f$ with a Multi-Layer Perceptron (MLP).
We make this choice because we want to approximate an arbitrary and nonlinear function, and we need a very small and simple model that can be trained quickly onboard the robots on the CPU.
Our fully-connected MLP has a single hidden layer of size 10 with a sigmoid activation function and L2 regularization of 0.01, and is trained online using RMSProp \cite{tieleman2012} as implemented in mlpack\footnote{https://mlpack.org/} with default parameters.
We normalize each input channel and perform a softmax on the output to obtain the final traversability probability.

The primary output of the local terrain analyser is the reachability scan $S(\phi)$, which is also associated with a pose.
$S(\phi)$ defines a star-shaped region that is known to be safe as well as the edges of obstacles.
Using the associated pose of $S$, we can label the traversability map pixels within the safe region as \texttt{trav} and the pixels just beyond the untraversable boundary \texttt{not\_trav}.
We assume that obstacles have a width of at least the size of a pixel of the obstacle map.
This assumption does not hold for thin obstacles like fences, but these obstacles are often not detected from the air, and in practice, we have nonetheless obtained good results.
We therefore now have a set of pixels with labels and feature vectors, and can train our MLP on this data, predicting traversability probabilities for all pixels on the map.
Training is done online on each robot while it is exploring, and is repeated every 10 seconds.
We perform no prior training and retain no data or weights between experiments, rather, all training data is gathered by the robot during the course of each experiment.

Given a graph edge $e = (e_1, e_2)$, we now want to compute its cost $C(e)$ for the purposes of graph search.
Recall that $C(e) =  -\log{P(e)}$, or the negative log of the probability of traversing the edge.
In order to approximate this, we discretize $e$ by sampling uniformly along the edge at points $s_{e,i}$ for $i \in [0, N_d]$.
Then
\begin{equation}
    C(e) = \sum_i -\log{T(s_{e, i})}
\end{equation}
Note that $C$ is the sum of negative log-likelihoods, meaning that when we solve for the lowest-cost path, we are maximizing the probability of traversal under the assumption that the traversabilities of all cells are independent.
This assumption is of course not true, but it does provide some physical intuition for the planner.
In practice, in order to handle pixels with unknown descriptors, we take the sum over the known points and then divide it by the number of known points, finally multiplying by $N_d$.

\begin{figure}
    \centering
    \includegraphics[width=0.3\textwidth]{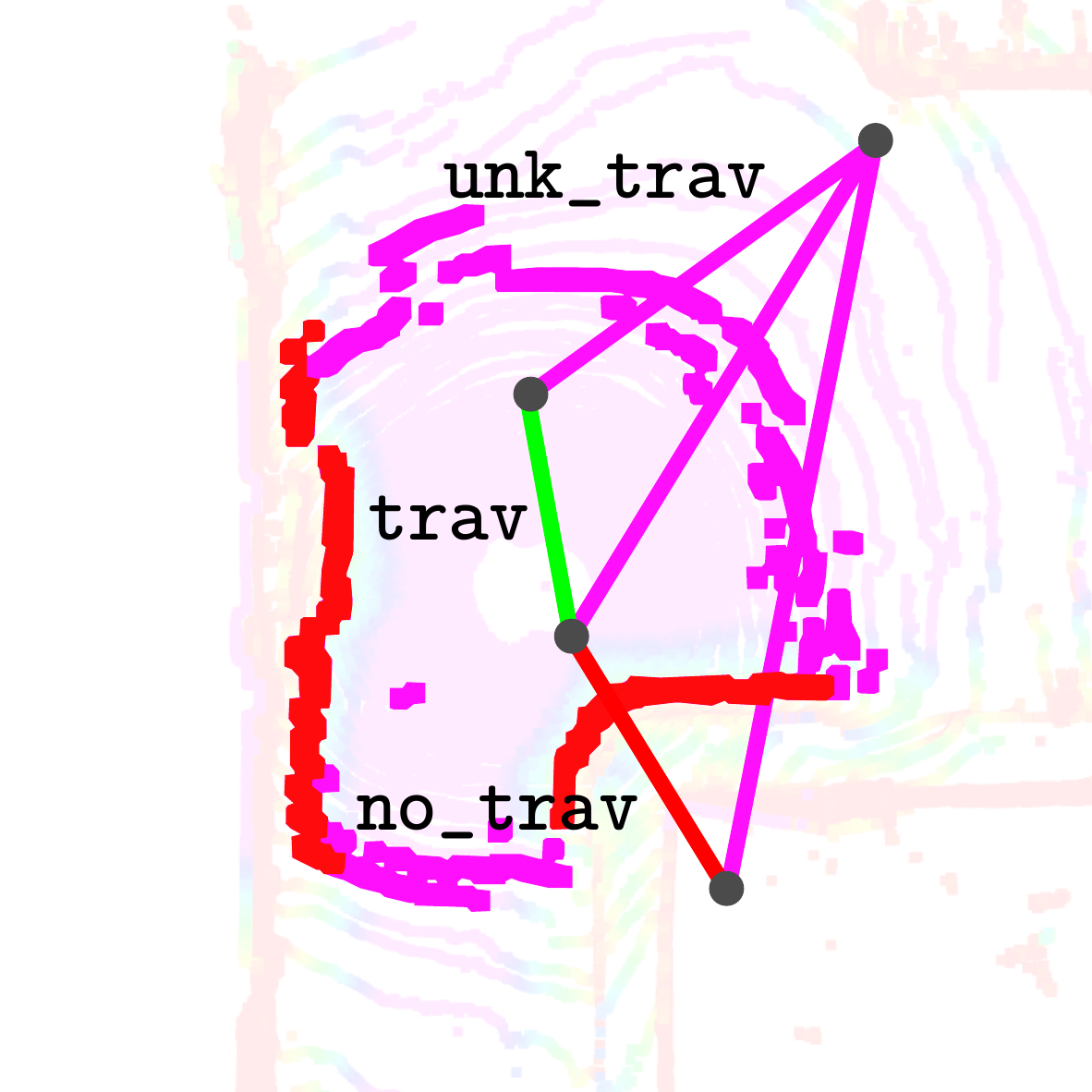}
    \caption{Reachability scan with several overlaid edges in the traversability graph labeled with their traversability determination.  The \texttt{trav} edge is considered traversable, \texttt{no\_trav} edge considered not traversable, and the \texttt{unk\_trav} edges considered unknown.}
    \label{fig:reach_edge_judge}
\end{figure}

\paragraph{Updating Graph with Experience:}
As we have observed, $S(\phi)$ defines a known safe region as well as known unsafe edges.
Given this information and an edge in the traversability graph, we would like to determine if the edge is traversable (\texttt{trav}), not traversable (\texttt{not\_trav}), or we are currently unable to determine the edge's traversability (\texttt{unk\_trav}).
These cases are illustrated in Fig.~\ref{fig:reach_edge_judge}.

Because the known safe region is defined as a star, we can compute a set of line segments between all the neighboring points.
We then check for intersections between this set of line segments and the query traversability edge.
If there are none and the endpoints lie within the safe region, then the edge must lie entirely in the safe region and therefore be \texttt{trav}.
Alternatively, if both edges lie outside the safe region and there are no intersections, the edge is \texttt{unk\_trav}.
If the edge has more than one intersection, we also assign it \texttt{unk\_trav}, adopting a wait-and-see approach.
If an edge has a single intersection, there are again two cases.
If there is an unknown point in $S$ within $\Delta \phi$ of the crossing, we assign \texttt{unk\_trav}, and if not we finally determine the edge to be \texttt{not\_trav}.

\begin{figure}
    \centering
    \includegraphics[width=0.8\textwidth]{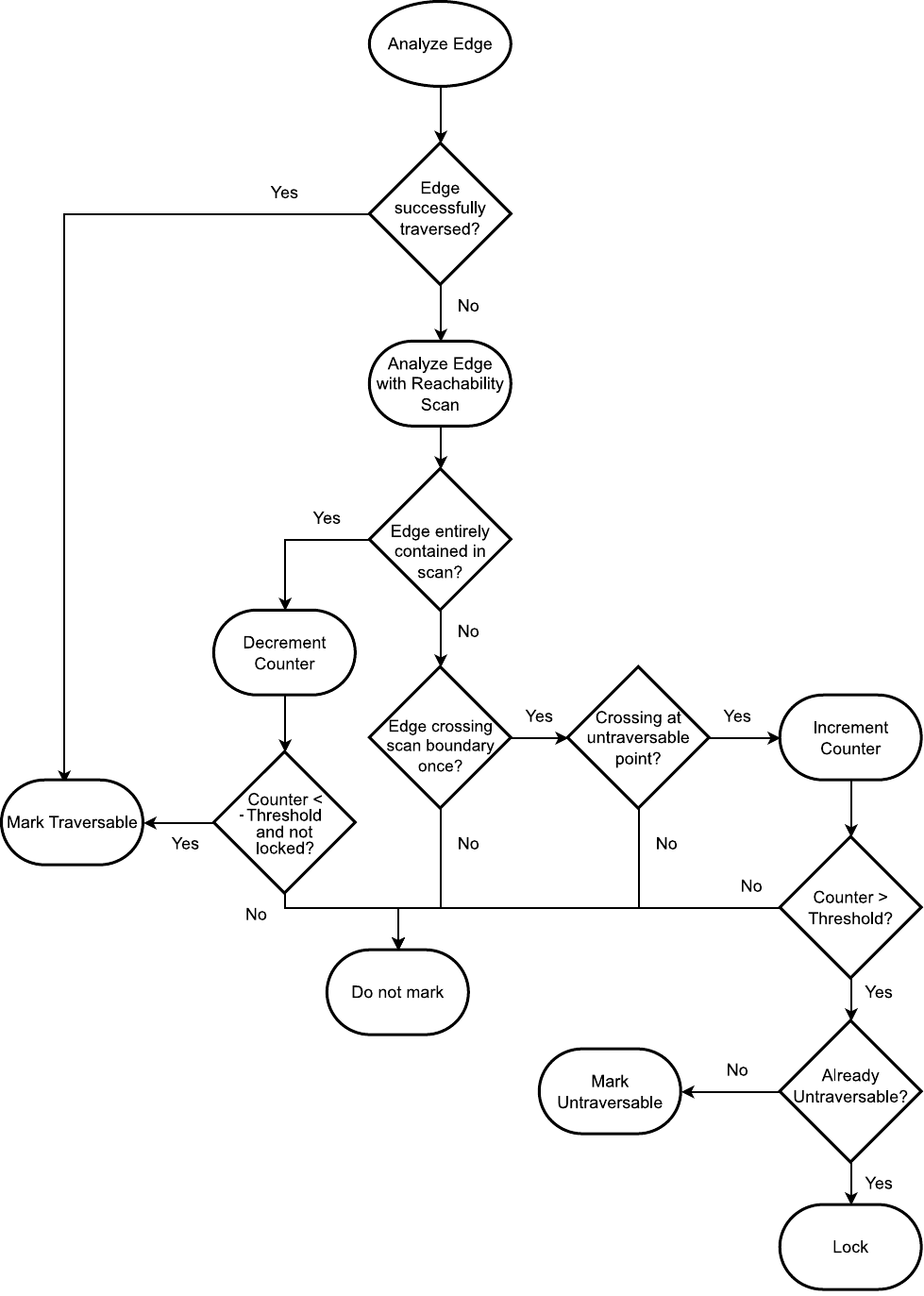}
    \caption{Flowchart showing the assignment of different edge states.}
    \label{fig:reach_exp_flowchart}
\end{figure}

If an edge is observed to be traversable, then we can assign it a very high traversability probability, and if it is not a very low one.
We override the estimated traversability from the map to do this.
In practice, observations are noisy, and it is important to not mark an edge as traversable if it is in fact not.
In this case, the robot may get stuck trying to traverse the edge and not realize that it is not possible to do so.
To address this, we require repeated, consistent observations to mark an edge.
Once an edge is marked \texttt{not\_trav} and \texttt{locked}, it can never be changed.
In addition, it cannot be traversed at all and is therefore effectively removed from the graph for the purposes of graph search.
An edge marked \texttt{trav} can be changed to \texttt{not\_trav}, but it requires even more consistent observations.
This procedure is described in more detail in Fig.~\ref{fig:reach_exp_flowchart}.

\paragraph{Sharing Experience:}
Using this framework, we now let robots share the accumulated sets of reachability scans that each have made.
All the robots are in the same coordinate frame due to the cross-view localizer, so robots can treat scans received by other robots in precisely the same way that they treat their own.
Scans are both incorporated into the map for the traversability model input as well as to directly update the graph.
We do prevent edges from being marked \texttt{not\_trav} on the basis of data from other robots.
In other words, the only way an edge can be completely removed from the graph is from a robots' own observations.
This prevents noisy or incorrect data from other robots from irrevocably affecting other robots' graphs.

\section{Simulation Experiments}
\begin{figure}
    \centering
    \includegraphics[width=\textwidth]{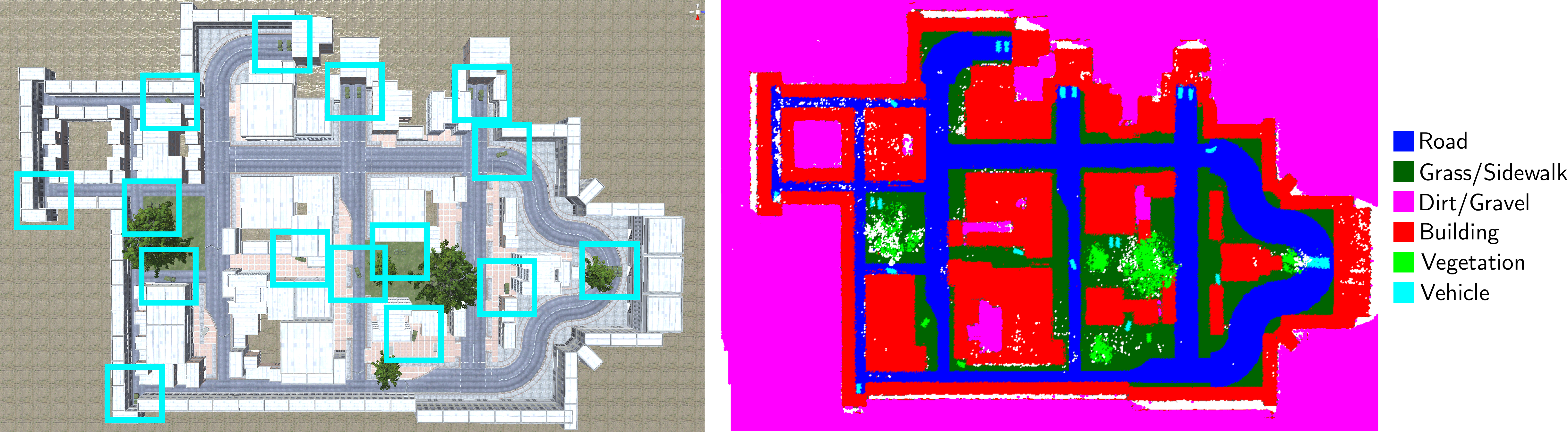}
    \caption{Simulation environment, with a color aerial image on the left where clusters of vehicles are highlighted with blue squares. The semantic map constructed by the UAV is on the right.}
    \label{fig:sim_city_env}
\end{figure}

We perform a number of experiments in our Unity-based simulator environment shown in Fig.~\ref{fig:sim_city_env}.
There are 15 clusters of cars in the environment, though in practice on occasion the goal detector will select multiple goals near one cluster.

For these experiments, we run the entire autonomy stack for all robots, with two main exceptions.
Firstly, semantic segmentation is simulated.
ASOOM is running, but just uses the ground truth image segmentation instead of ERFNet.
Secondly, instead of running LLOL, we use ground truth odometry with a small amount of Gaussian noise added.
We additionally simulate the LLOL depth panoramas and their semantically segmented counterparts every second, which are fed into the downstream autonomy stack.
We make these choices for several reasons:
\begin{itemize}
    \item Segmentation and odometry are the most computationally expensive portions of the system, so simulating them allows us to test larger teams.
    \item Segmentation and LiDAR odometry are very well-studied algorithms, and not the focus of this work.
    \item High-fidelity sensor simulation for reasonable segmentation and odometry testing is difficult to obtain, and without it, running these algorithms in the loop is not representative of real-world performance.
\end{itemize}

The distributed communication system is also simulated.
If two robots come within 50 meters of each other, they synchronize their last known positions.
If they come within 30 meters, they synchronize everything.
This approximates the behavior of the distributed communication system where the highest priority messages may start syncing first, and after some time or after the link quality improves the full synchronization is completed.
There is a basestation at the starting point, which like the basestation in the real world experiments acts as a static node for the distributed communication system that all robots synchronize with opportunistically.
All experiments are run for 30 minutes, and we run 10 copies of the same experiment for each configuration.

\subsection{Scaling}
\begin{figure}
    \centering
    \begin{subfigure}
        \centering
        \includegraphics[width=0.4\textwidth]{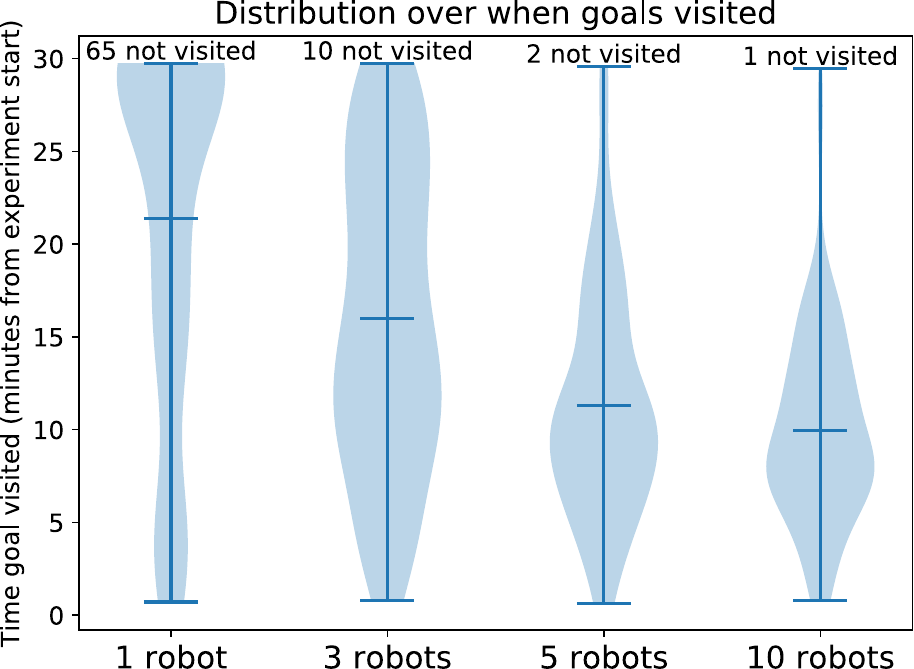}
    \end{subfigure}
    \begin{subfigure}
        \centering
        \includegraphics[width=0.4\textwidth]{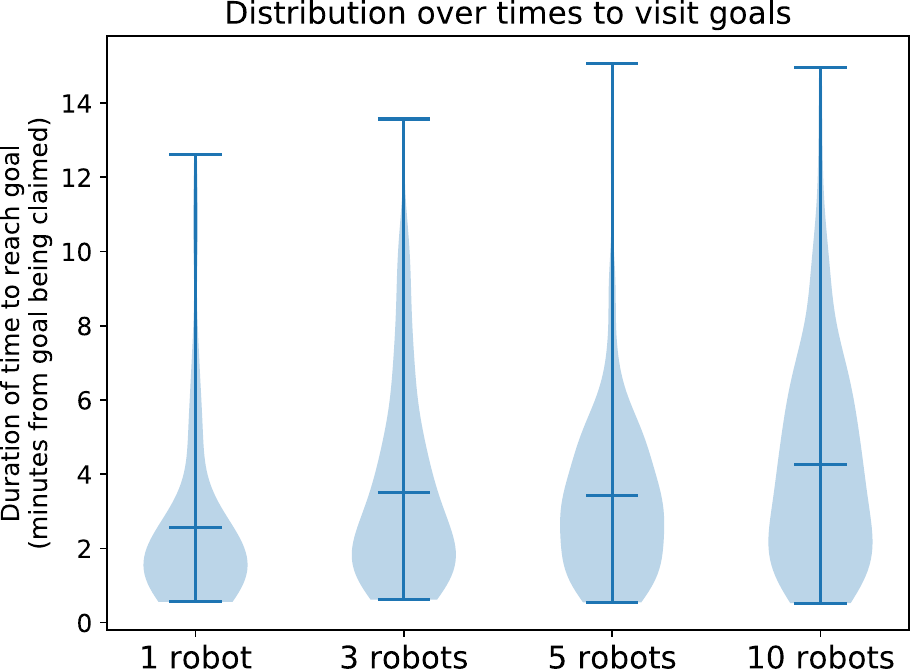}
    \end{subfigure}%
    \caption{(Left): Distribution of the time taken to visit goals after mission start.  Goals not visited are assigned a default visit time of 30 minutes for the purposes of visualization. (Right): Distribution of the time it takes between a given robot claiming and reaching a goal, for those goals which are visited (see Fig.~\ref{fig:coord_state_machine})}.
    \label{fig:scale_visit_dists}
\end{figure}

We first investigate the performance of our system as the number of UGVs is increased.
In all of these experiments, we have a single UAV.
In Fig.~\ref{fig:scale_visit_dists} we visualize the times that goals are visited.
Note that as the number of UGVs increases, the goals are visited earlier, and more goals are visited, suggesting that our system is able to utilize multiple robots to improve overall performance.
Interestingly, increasing the number of robots also appears to increase the time taken to reach a goal, but this is likely mainly due to the further goals not being reached with fewer robots, skewing the distribution.

\begin{figure}
    \centering
    \includegraphics[width=0.5\textwidth]{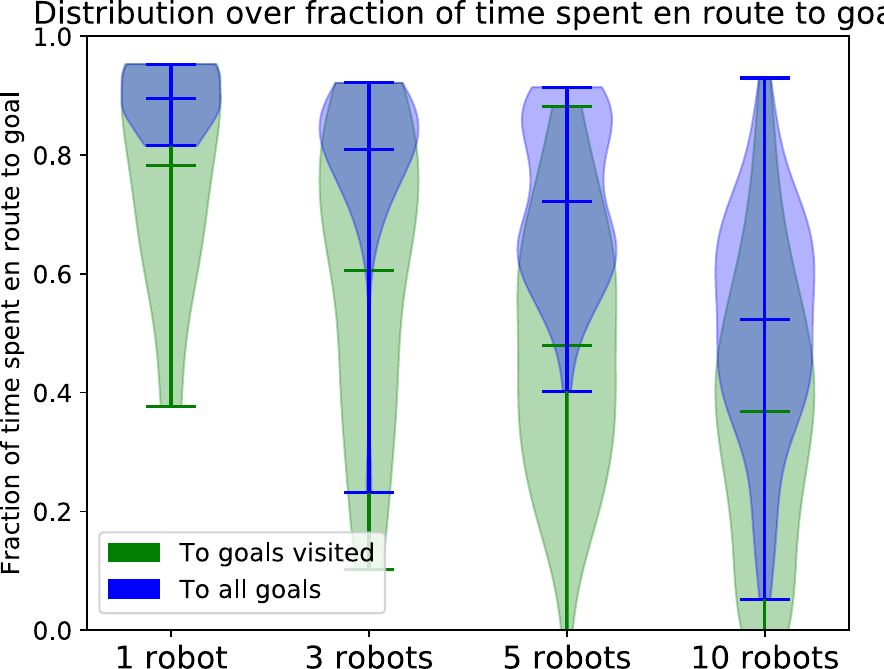}
    \caption{Distributions of the fraction of time robots spend actively navigating to a goal for different team sizes.}
    \label{fig:scale_timespentdist}
\end{figure}

We also analyze how the robots are spending their time.
Robots may be either headed to a goal or not.
If they are headed to a goal, they may ultimately visit that goal, or they may time out or be preempted.
In Fig.~\ref{fig:scale_timespentdist} we visualize the distribution over all robots and experiments.
Note that the fraction of time that a robot spends headed to a goal decreases as the number of robots goes up.
This suggests that while adding more robots helps, there are also diminishing returns because those robots spend less time doing useful work.

\begin{figure}
    \centering
    \begin{subfigure}
        \centering
        \includegraphics[width=0.4\textwidth]{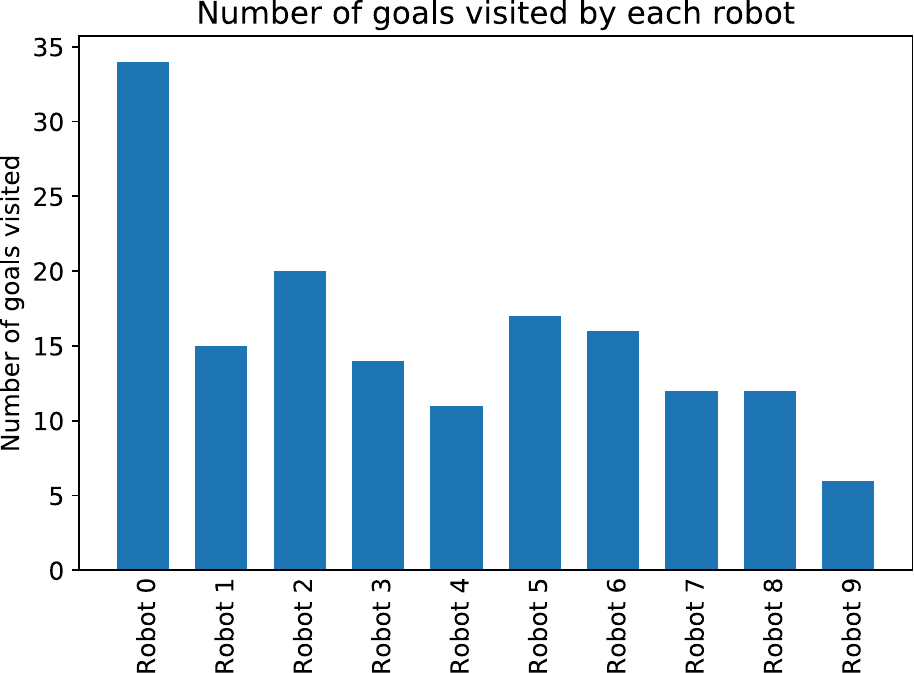}
    \end{subfigure}
    \begin{subfigure}
        \centering
        \includegraphics[width=0.4\textwidth]{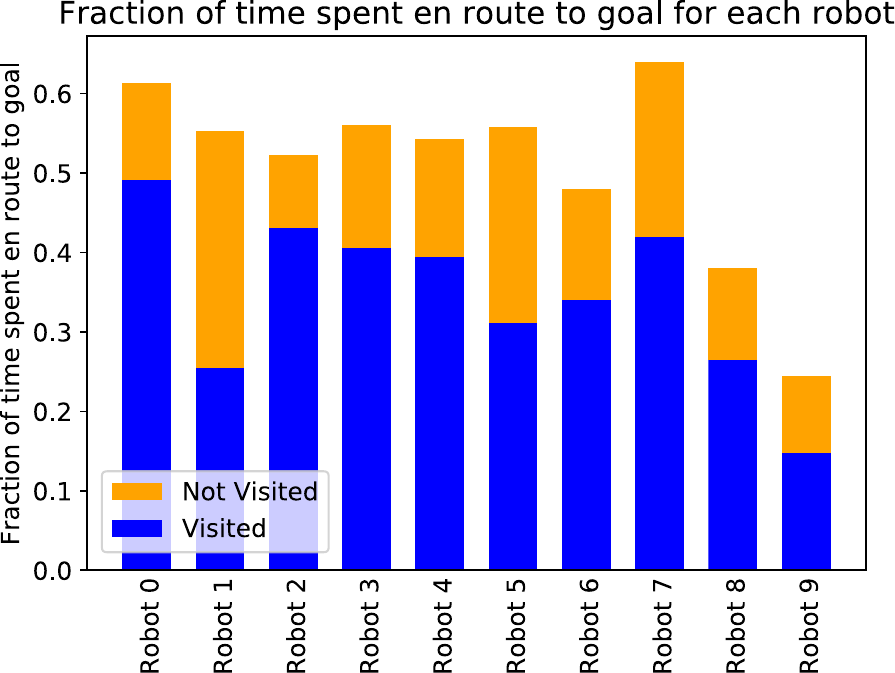}
    \end{subfigure}%
    \caption{Robots' productivity on a per-robot basis for a 10-UGV team.}
    \label{fig:10robots_timespent}
\end{figure}

We further analyze this phenomenon on a per-robot basis in Fig.~\ref{fig:10robots_timespent}.
Robots are organized in decreasing priority, and note that robots of lower priority generally visit fewer goals as well as spend less time visiting goals.
This suggests that there simply are not enough goals being fed to all robots to use them all effectively.
Given that there are approximately 15 goals, 10 robots, and the goals are gradually discovered over time, this is an intuitive result.

All of these experiments are considering only a single UAV.
We propose two ways of scaling up the number of UAVs.
In our prior work \cite{cladera2023enabling}, two UAVs are deployed, where the mission of one UAV is to perform only communications.
This approach may lead to network congestion with most data travelling through a single node, decreasing the overall performance of the system.
We can solve this problem by running multiple parallel teams, where each team independently explores a different area.
This scales in a perfectly parallel fashion, avoiding network congestion, but we lose any benefits of interaction between teams beyond simply dividing up the environment.
Under this architecture, the relevant scaling metric is the UAV/UGV ratio, as each UAV is assigned an independent UGV team.
However, if inter-team communication and collaboration are allowed, the system becomes more complex, and we leave this for future work.

\subsection{Learning Traversability}
In order to analyze the effectiveness of our terrain analysis and learning system, we perform several ablation experiments.
For these experiments, we use a team size of 5 robots.
In one set of tests, we disable the terrain learning system.
Instead, we rank the classes by traversability and assign each edge a number $T_{cls}$ corresponding to the most difficult class to traverse.
We then weight the edge $W = 100^{T_{cls}} - 1 + D$, where $D$ is the minimum distance from a point on the edge to the nearest next-highest class.
This causes the planner to find a path entirely on the easiest class, and if that fails to include the next highest class, and so on.

We perform another set of tests where in addition to disabling the learning system, we also disable any updates of the traversability graph.
In other words, the traversability graph is frozen from the beginning, and there is no global replanning at all.
All global paths planned are therefore static.

\begin{figure}
    \centering
    \begin{subfigure}
        \centering
        \includegraphics[width=0.4\textwidth]{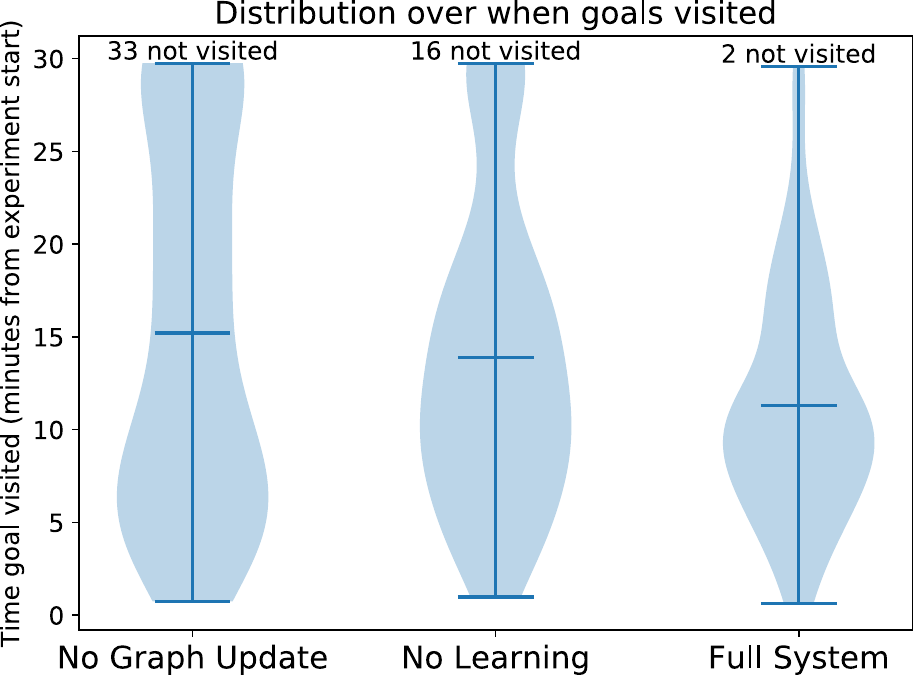}
    \end{subfigure}
    \begin{subfigure}
        \centering
        \includegraphics[width=0.4\textwidth]{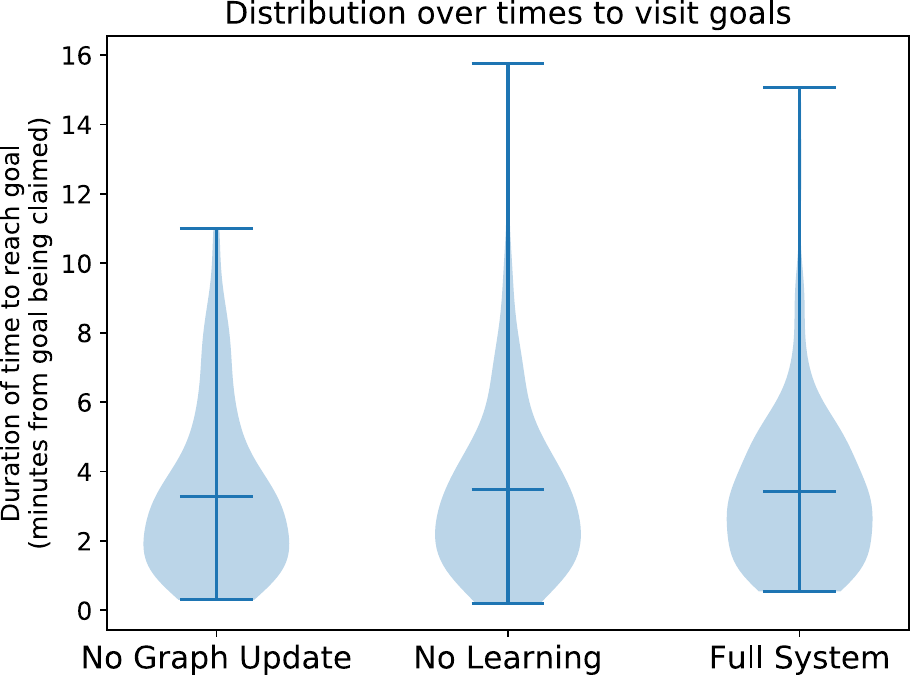}
    \end{subfigure}%
    \caption{For different terrain learning ablations, (Left): Distribution of the time goals are visited from mission start.  Goals not visited are assigned a visit time of 30 minutes for the purposes of visualization. (Right): Distribution of the time it takes a given robot to reach a goal once that robot has claimed the goal, for those goals which are visited.}
    \label{fig:terrainlearn_visit_dists}
\end{figure}

In Fig.~\ref{fig:terrainlearn_visit_dists} we perform the same analyses as in Fig.~\ref{fig:scale_visit_dists}, but instead with varying terrain learning modes.
Note that removing updates to the graph (that is, disabling global replanning) significantly affects the team's performance.
It does not have a large effect on the time to visit goals that are eventually reached, but it does decrease the number of goals that are ultimately visited.
For all experiments, we erect several roadblocks in the environment which are labeled semantically as road.
Without global replanning, the robots are unable to handle these obstacles, because they have no way of adapting their global plan.

\begin{figure}
    \centering
    \includegraphics[width=0.65\textwidth]{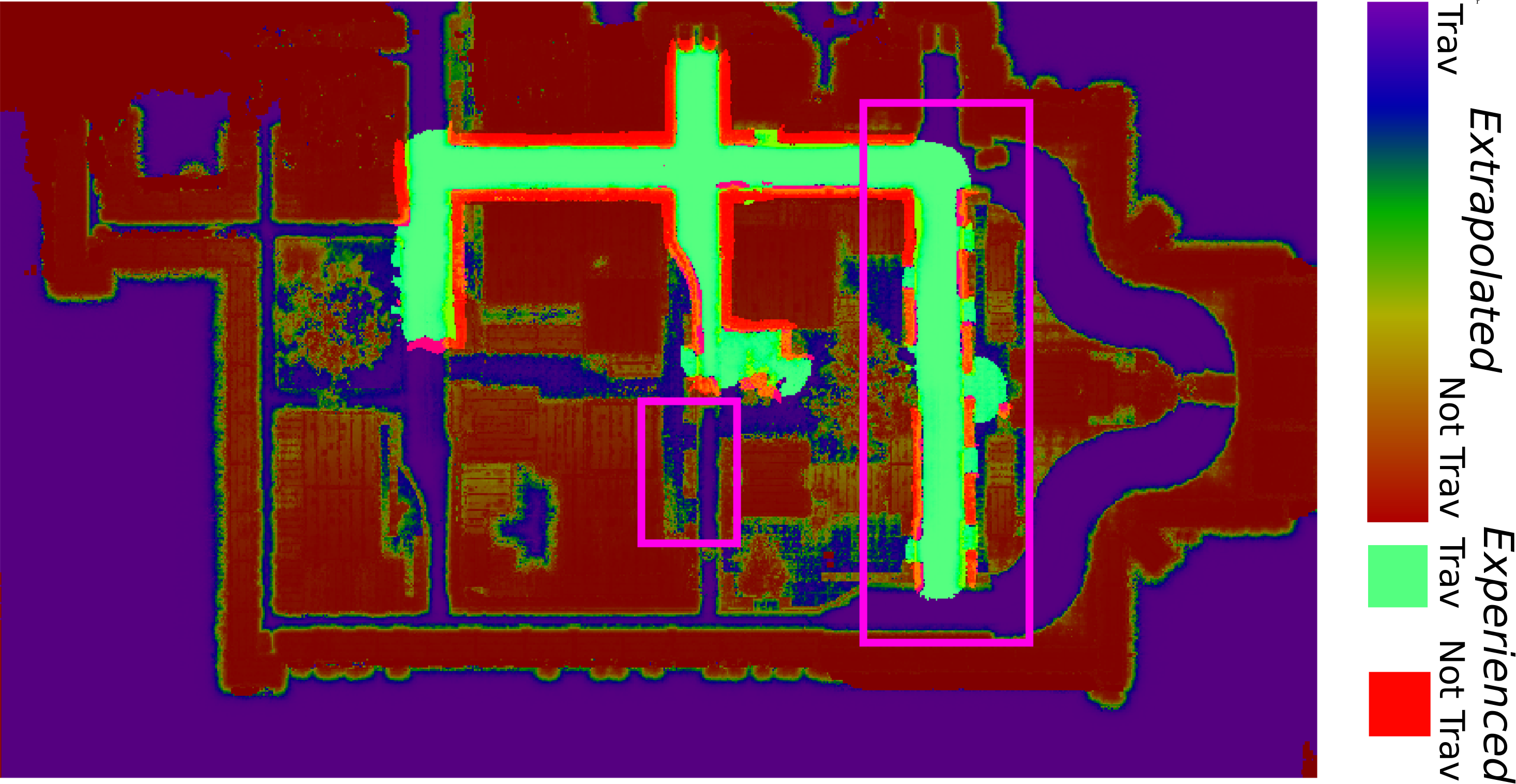}
    \caption{Visualization of the terrain analysis partway through a simulated experiment.  The pink boxes highlight regions where the robots observed ramps up the curb, and areas where these ramp detections were extrapolated to regions seen only by the UAV.}
    \label{fig:learned_trav_example_sim}
\end{figure}

Removing the terrain learning system also has an effect.
Fig.~\ref{fig:learned_trav_example_sim} gives us some clues as to why.
The environment has a number of curbs between the sidewalks and roads.
There are ramps that the robots can use to drive up onto the sidewalk from the roads, but these have the same semantic class as sidewalk.
Some of these areas are highlighted by pink boxes in Fig.~\ref{fig:learned_trav_example_sim}.
Note that the robot experiences ramps in one area of the map but is able to extrapolate them to a different region.
Even though they are the same semantic class, the model is able to learn cues from the ramp colors.
This means that for those goals which are on the sidewalks, robots are able to find paths up ramps to them much more quickly than without the terrain learning system.

\begin{figure}
    \centering
    \includegraphics[width=0.45\textwidth]{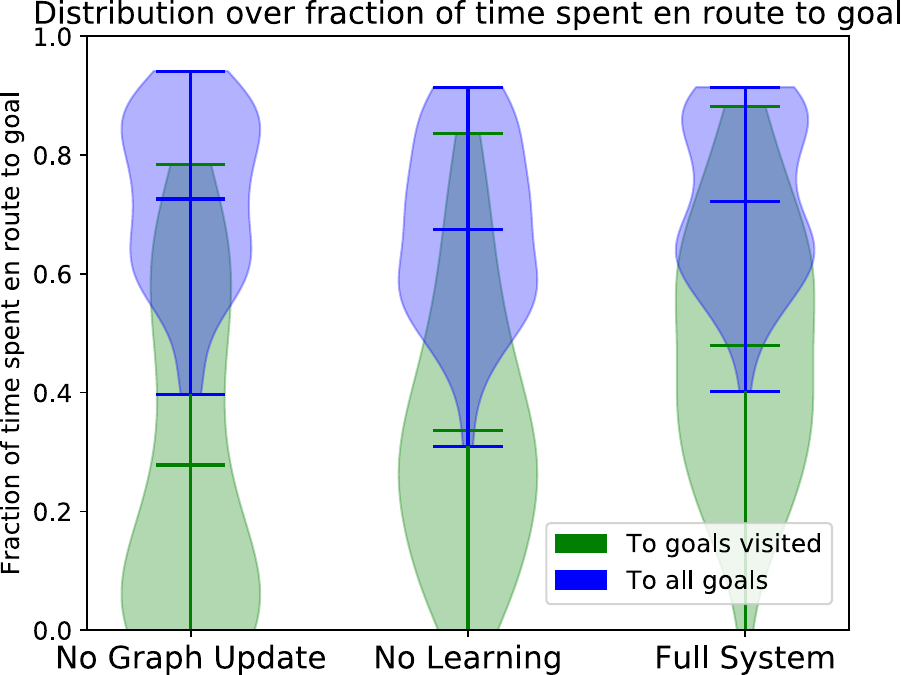}
    \caption{Distributions of the fraction of time robots spend actively navigating to a goal for different terrain learning ablations}
    \label{fig:terrainlearn_timespentdist}
\end{figure}

We additionally analyze the robots' time as in Fig.~\ref{fig:scale_timespentdist} in Fig.~\ref{fig:terrainlearn_timespentdist}.
The effect here is less pronounced, which is unsurprising.
Robots spend about the same amount of time going somewhere, but the full system spends a larger time going to a goal that they will eventually visit.
This is due to the goals being reached instead of timing out, and correlates with the reduced number of goals that are never visited.

\subsection{Communication}
\begin{figure}
    \centering
    \includegraphics[width=0.5\textwidth]{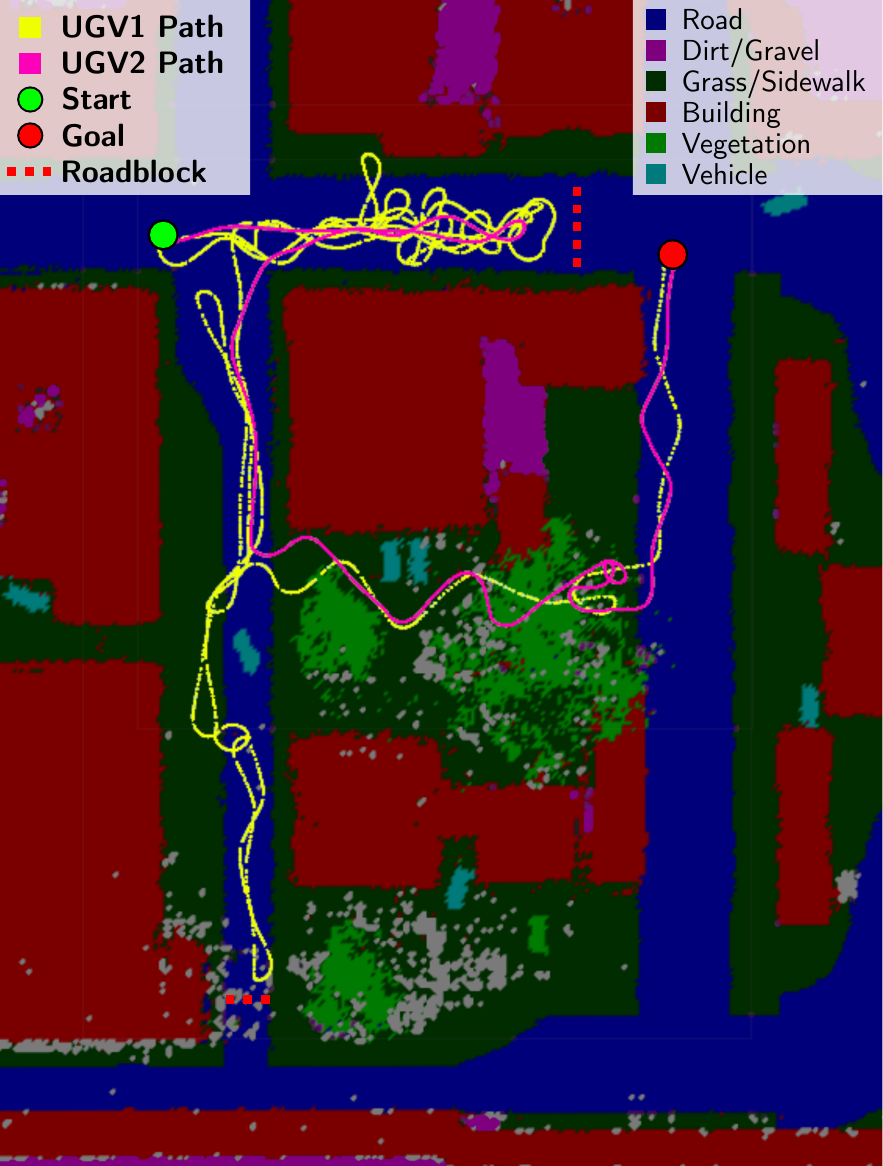}
    \caption{Trajectories of two robots navigating to the same goal.  UGV 2 is given UGV 1's experience before beginning.}
    \label{fig:robot_comm_demo}
\end{figure}

Along with ablating how the robot uses the terrain data, we also ablate how the robots share terrain information with each other.
To demonstrate the utility of this, we first perform a simple experiment.
Two robots are given the same map, and one is asked to navigate to a target position.
Once it arrives there and the other robot has gotten all of the other robot's reachability scans, that other robot is given the same goal.
The results of this experiment can be seen in Fig.~\ref{fig:robot_comm_demo}.
Notably, there are several roadblocks on the map, including one directly between the start and goal positions.
The first robot must therefore explore several possible routes which turn out to not be traversable before finding a route.
The second robot, by contrast, is able to use the first robot's experience to much more quickly find the correct route.
It does not trust the first robot's data without question and does perform some exploring on its own, but overall is far more efficient by virtue of the other data.
We compare the two trajectories in Table \ref{tab:robot_comm_demo_comp}.
Note that UGV 2 is more than 3 times faster and takes nearly 3 times less distance to reach the same goal, since it is able to leverage UGV 1's experience instead of taking the time to rediscover the same things.

\begin{table}
    \centering
    \begin{tabular}{c|c|c}
        & Distance (km) & Time (min) \\
        \hline
        \hline
        UGV 1 & 0.88 & 20.6 \\
        \hline
        UGV 2 & 0.30 & 6.4
    \end{tabular}
    \caption{Two robots' times to a goal, where UGV 2 is given UGV 1's experience.}
    \label{tab:robot_comm_demo_comp}
\end{table}

\begin{figure}
    \centering
    \begin{subfigure}
        \centering
        \includegraphics[width=0.4\textwidth]{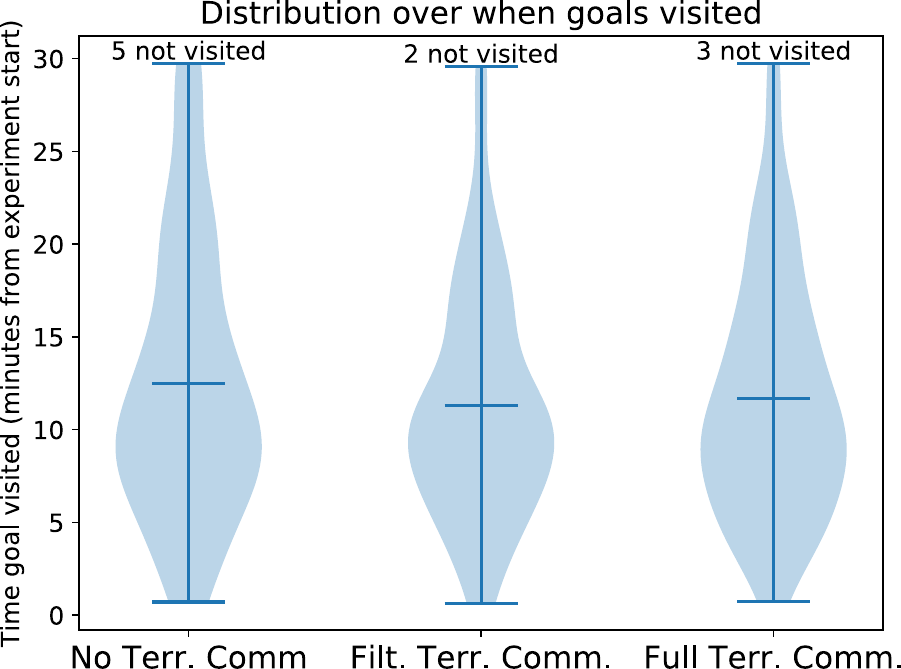}
    \end{subfigure}
    \begin{subfigure}
        \centering
        \includegraphics[width=0.4\textwidth]{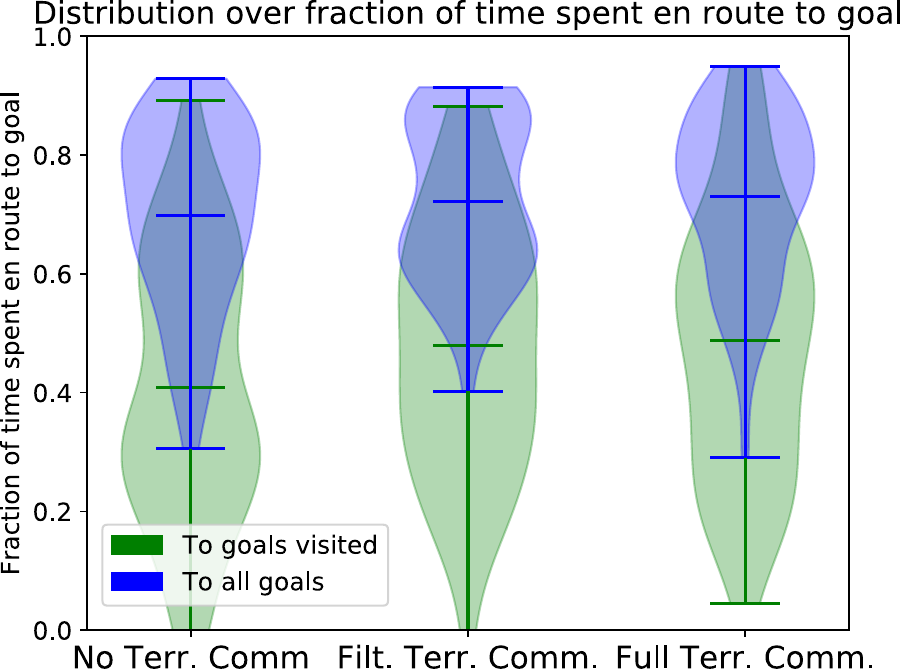}
    \end{subfigure}%
    \caption{For varying communication, (Left): Distribution of the time goals are visited from mission start.  Goals not visited are given a visit time of 30 minutes for the purposes of visualization. (Right): Distributions of the fraction of time robots spend actively navigating to a goal.}
    \label{fig:terraincomm_analysis}
\end{figure}

In addition to this toy experiment, we also run large scale experiments, adjusting how much terrain information is shared.
In one set of experiments we turn off communications about terrain completely, though robots do still share information about their location and target goals for coordination.
In another set the robots share all of their reachability scans, so in effect each robot has the same total information, assuming all robots are synchronized.
Finally, we perform a set of experiments where the robots share a downsampled set of reachability scans.
Each robot checks to see if a new reachability scan is at least 2 meters away from all prior scans.
If it is, it is shared with other robots.
This is the approach used in all other experiments, including those in the real world.
These results are shown in Fig.~\ref{fig:terraincomm_analysis}.
Sharing traversability information does appear to result in slightly increased team effectiveness, with robots visiting goals earlier and spending more time navigating to a goal they will ultimately visit.
However, sharing the full information has minimal effect, and even slightly increases the average time goals are visited.
Qualitatively, the differences are most notable at the very beginning of the experiment.
After all the robots have traveled a little bit, their inferred traversability maps are fairly decent.
Therefore, while sharing information helps the traversability estimation to converge to something reasonable faster, after it has done so the effect is relatively small, at least for this particular environment.

While it is possible that the small increase in average time goals are visited with full communication is due to statistical variations, we did also observe qualitatively one reason it could happen.
Sometimes, a robot determines a path to be not traversable when it actually is.
This is by design, since the system is tuned to be conservative, as it is better in most cases to be cautious than risk damage.
However, if all information is shared and a robot makes a poor determination like this, then all other robots will more easily make that same mistake.
Downsampling helps to regularize other robots' measurements, resulting in robots being more willing to explore regions others have not been able to.
In summary, sharing traversability information is most valuable when robots are exploring the same region sequentially, and less helpful in simultaneous operation.

We note as well that our system fails gracefully as communication degrades.
In other words, the task can be executed, even if no communication about terrain can occur at all.
However, all of these experiments do assume some level of communication.
Our system is dependent on the UAV being able to communicate the orthomap to the UGVs.
If communication is so degraded, or the map so large, that this cannot happen, it will fail, though we could instead provide a static map \textit{a priori}.
In addition, UGVs must be able to communicate to deconflict goals.
In the absence of communication, robots could simply pick goals at random, but this would result in significant redundant goal visits.
We do assume a basic level of communication to be available, but anything above and beyond that, while helpful, is not strictly necessary.

\section{Real World Experiments}
Finally, we perform experiments in the real world.
We emphasize that these were not controlled environments.
Cars were frequently going by, construction was ongoing with roadblocks and construction equipment, and there was a great deal of complicated structure.
We test in two environments: Pennovation and the New Bolton Center (NBC).
In all cases, our team consisted of 3 Jackal UGVs and a single UAV.
The Jackals were equipped with an AMD Ryzen 3600 CPU, NVidia GTX 1650 GPU, Ouster OS1-64 LiDAR, and Realsense D435i camera (only used for visualization).
The UAV was a custom PX4-based robot \cite{liu2022large}, equipped with an Intel NUC i7-10710U, UBlox ZED-F9P GPS, and Open Vision Computer (OVC) \cite{quigley2019open} global shutter RGB camera (we did not use the stereo pair or IMU).
All robots used 5 GHz Rajant Breadcrumb DX2 mesh radios, with a Rajant ME4 basestation.

\begin{figure}
    \centering
    \includegraphics[width=0.6\textwidth]{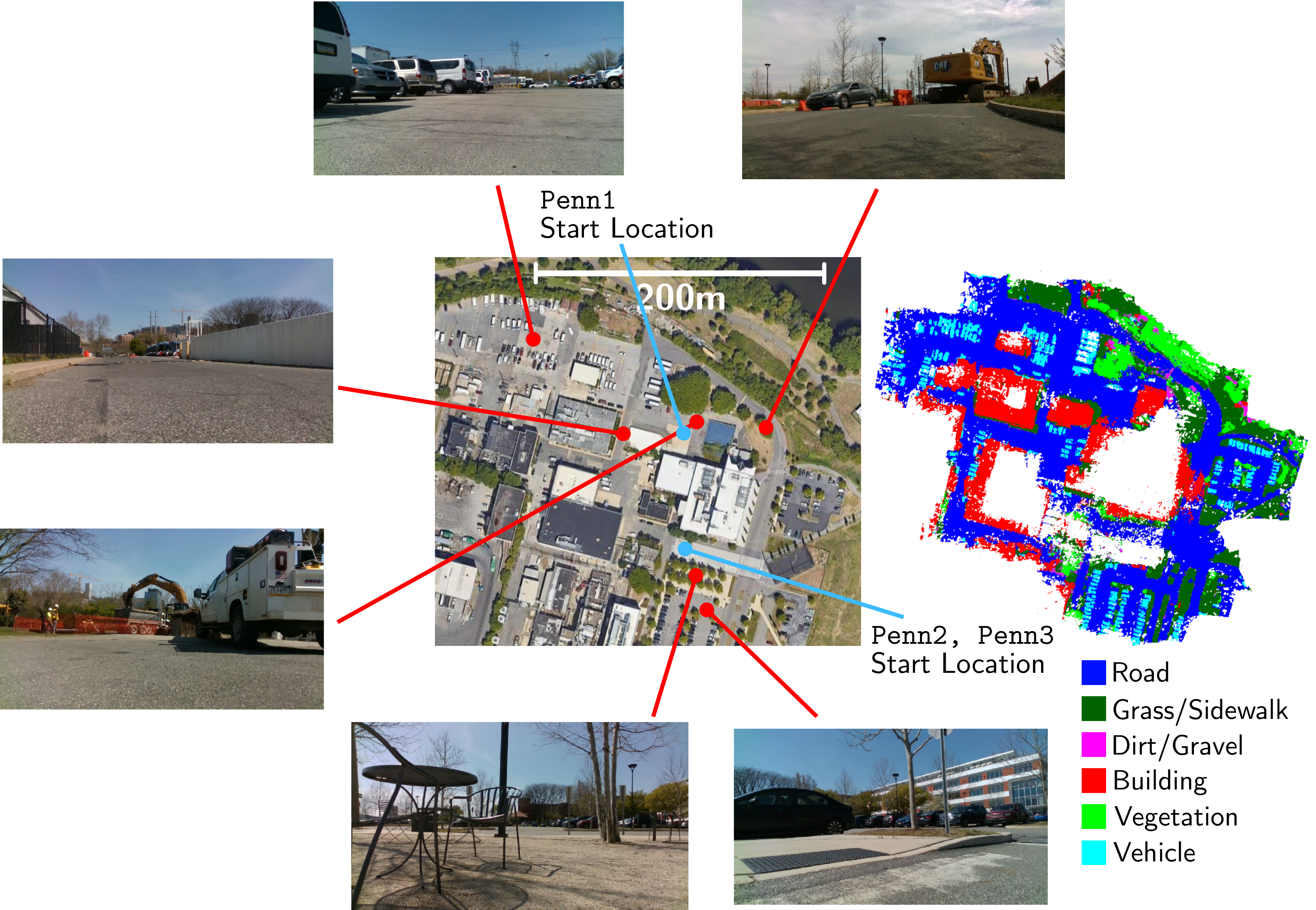}
    \caption{Satellite and semantic map of the Pennovation environment, with some robot's-eye views of some representative areas.}
    \label{fig:pennov_env}
\end{figure}

An overview of the Pennovation environment is shown in Fig.~\ref{fig:pennov_env}.
This is a former factory refurbished to accommodate a combination of labs, offices, and storage.
It offers a relatively urban setting, though there are some green spaces and trees.
Particularly challenging regions include a cluttered patio with poles, tables, and chairs, as well as parking lots which contain many curbs and ramps.
During our experiments there was ongoing road construction, resulting in roadblocks, construction equipment, and increased traffic due to the main road being reduced to one lane.
We perform 3 experiments here, with \texttt{Penn1} starting the UAV and UGVs in a different area of the map from \texttt{Penn2} and \texttt{Penn3}, and therefore using a different exploration path for the UAV.
The basestation in all cases was positioned at the start location.

\begin{figure}
    \centering
    \includegraphics[width=0.6\textwidth]{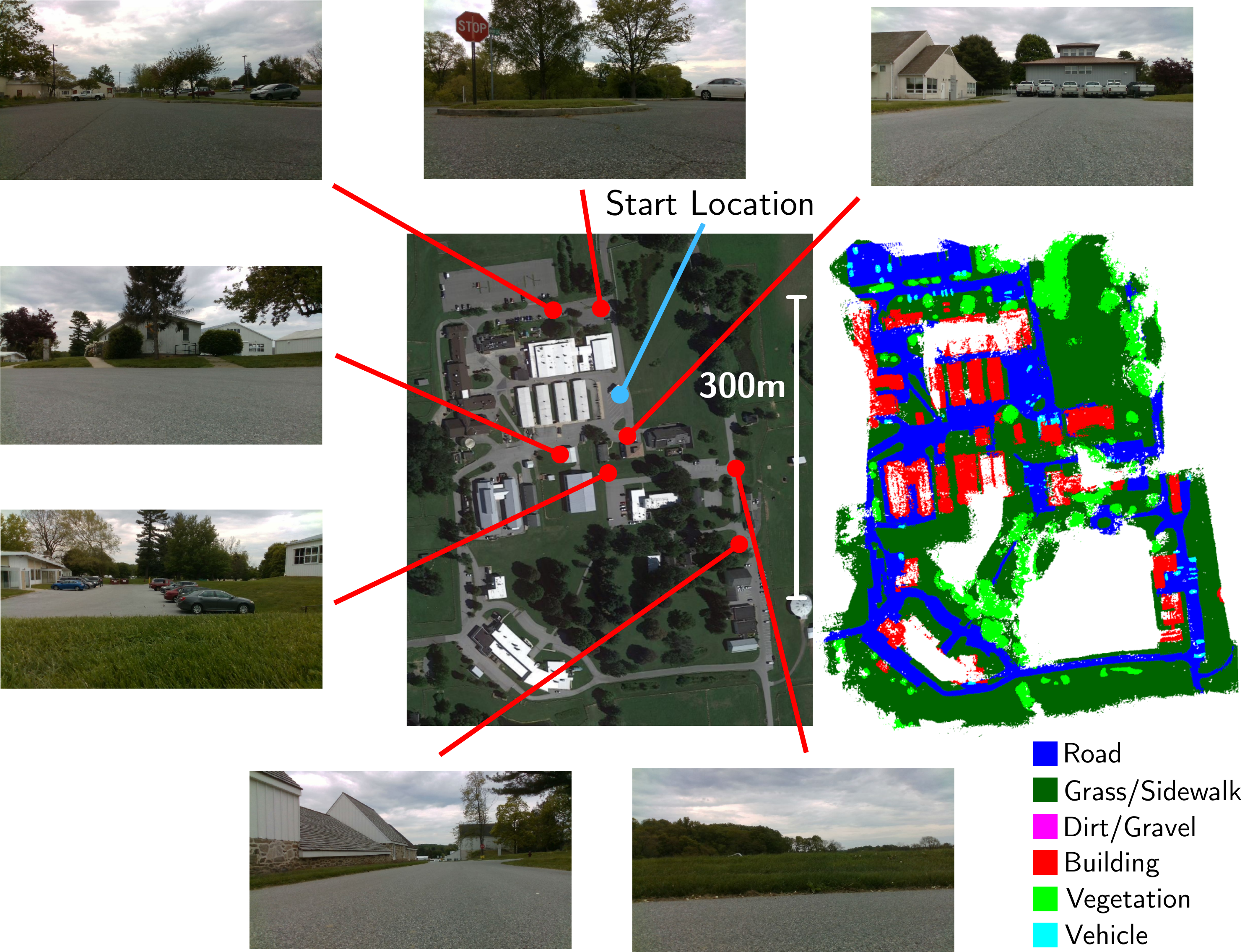}
    \caption{Satellite and semantic map of the NBC environment, with some robot's-eye views of some representative areas.}
    \label{fig:nbc_env}
\end{figure}

An overview of the NBC environment is shown in Fig.~\ref{fig:nbc_env}.
For this experiment we did not fly the UAV in real-time due to a hardware failure of the aerial robot.
Instead, we fed the UGV a complete aerial map that was built earlier by ASOOM at the beginning of their mission.
Robots were able to communicate with each-other and the basestation directly when in range, but not through the UAV in this case.
We are therefore unable to evaluate the UAV's effect on this experiment, but nonetheless do still exercise the coordination and traversability-sharing parts of the system.
This environment is somewhat larger scale and much more rural, consisting of roads between scattered buildings.
There are some more significant elevation changes here as well as many more trees and much more grass compared to Pennovation.

Table \ref{tab:exp_summary} summarizes our 4 experiments.
We ended an experiment either when all robots returned back to the start point with all identified goals visited, or after the UAV battery died and it appeared that not all goals would be visited soon.
Over all experiments, we visited 49/51 goals, or 96\%.
Both times a goal was not reached the goal was in a similar location, which was in a parking lot very challenging to reach due to construction on the access road to it.
In \texttt{Penn1} it was particularly difficult because the UAV misclassified a portion of the road as building, causing the global planner to struggle.
In \texttt{Penn2} the same goal was almost reached, but the goal timed out just before the robot got there.
UGVs traveled a total of 17.8km autonomously across all the UGVs and experiments, constituting a total of 4.2 UGV-hours.
Each individual UGV had on average 1.4 hours of autonomous operation and traveled 5.9km.

\begin{table}
    \centering
    \small
    \begin{tabular}{c||c|c|c}
         Dataset & Duration (min) & Goals Visited/Total Goals & Total Dist Travelled (km) \\
         \hline
         Penn1 & 28.5 & 10/11 & 3.9 \\
         Penn2 & 28.9 & 15/16 & 4.8 \\
         Penn3 & 28.0 & 16/16 & 5.0 \\
         NBC1 & 31.4 & 8/8 & 4.1
    \end{tabular}
    \caption{Overview of the scale of 4 full-scale real-world experiments in terms of duration, goal count, and distance travelled.}
    \label{tab:exp_summary}
\end{table}

During our experiments, each robot was followed by a safety driver with the ability to take over manually if need be.
In Table \ref{tab:takeover_analysis} we analyse the robustness of the autonomy system by looking at the fraction of time each robot was taken over manually.
A large portion of these takeovers were due to dynamic obstacles, which our planner was not designed to handle.
Most often these obstacles were vehicles since our experiments were conducted on active roads and parking lots.
In addition, on occasion, the robots had to be prevented from colliding with each other.
Finally, there were several speed bumps in the Pennovation environment, and the robots were manually driven over these to reduce mechanical wear and tear on the robots.
Because dynamic obstacles and speed bumps were cases our system was not designed to handle, we refer to the fraction of time spent in manual mode, ignoring these cases, as \textit{filtered} in Table \ref{tab:takeover_analysis}.

Most of the time, robots spent only 1-3 \% of the time in manual mode, corresponding to on the order of 5 takeovers per robot per experiment.
These unfiltered takeovers were done to avoid collisions at the discretion of the human safety driver.
The most common cause was curbs, particularly when approached from above with the curb as a negative obstacle.
In this case, it was difficult for the robot to distinguish a curb from a relatively mild downslope.
Other challenging obstacles included poles and other small objects like chairs.
Due to the polar terrain analysis, the inflation radius of objects very close to the robot tends to be inaccurate, sometimes causing the local planner to drive into nearby narrow obstacles.
\texttt{Penn1} had a significantly higher number of takeovers due to the experiment starting next to a parking lot with many curbs and a patio with many trees, poles, and chairs.
Finally, on some rare occasions, the robots tried to drive under overhangs, such as large trucks and construction equipment.
While it is possible that they would have been able to do so successfully, the operator intervened out of an abundance of caution.
It is also worth noting that for \texttt{NBC1}, one UGV tried to drive over a particularly steep rut next to the road, and while it made it over, the impact caused the robot to shut down.

\begin{table}
    \centering
    \small
    \begin{tabular}{c||c|c|c|c|c}
         Dataset &
         \begin{tabular}{c}
              UGV1 \% manual \\
              (raw/filtered)
         \end{tabular} & UGV2 \% manual & UGV3 \% manual & Overall \% manual \\
         \hline
         Penn1 & 32.2/23.1 & 8.1/3.9 & 15.7/11.8 & 17.9/11.9\\
         Penn2 & 5.9/3.0 & 3.8/0.07 & 3.2/0.36 & 4.5/1.3 \\
         Penn3 & 18.8/7.6 & 7.7/1.0 & 9.7/0.81 & 12.0/3.0 \\
         NBC1 & 3.9/1.8 & 3.4/3.2 & 3.7/0 & 3.7/1.8
    \end{tabular}
    \caption{Analysis of the fraction of time robots spent being driven manually across different real-world experiments.}
    \label{tab:takeover_analysis}
\end{table}

\begin{figure}
    \centering
    \includegraphics[width=0.7\textwidth]{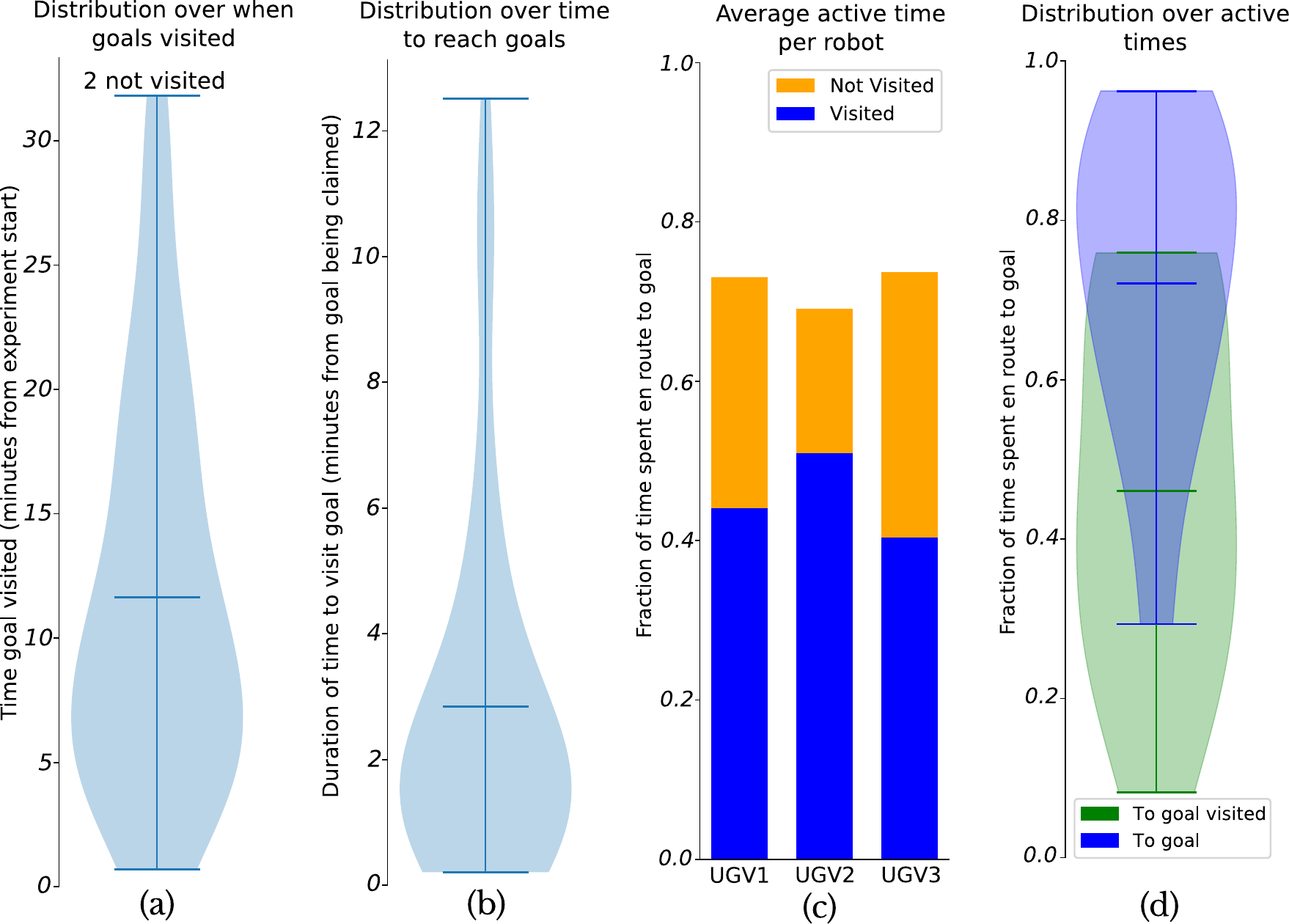}
    \caption{Analysis of goal timings and robot activity throughout the real-world experiments.}
    \label{fig:real_exp}
\end{figure}

During all experiments, the communication system and UAV planner performed well.
It was rare that UGVs did not have any active goals to visit, as can be seen in Fig.~\ref{fig:real_exp} (c) and (d).
We can also see in Fig.~\ref{fig:real_exp} (a) that the majority of the goals were usually visited in the first half of the experiment, with the long tail at the end being often due to particularly challenging or distant goals to visit.

\begin{figure}
    \centering
    \includegraphics[width=0.9\textwidth]{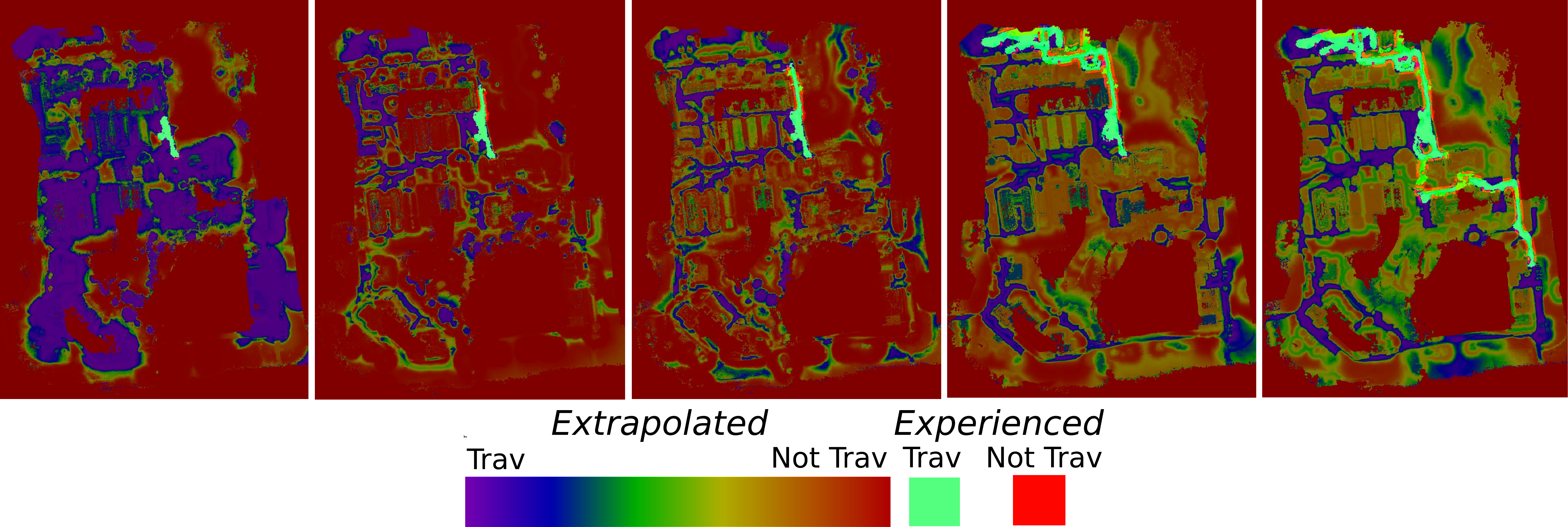}
    \caption{The learned terrain model and experienced terrain for one of the UGVs over the course of the \texttt{NBC1} experiment.}
    \label{fig:nbc_learn_trav}
\end{figure}

We also analyse the learned terrain model from \texttt{NBC1} over the course of the experiment in Fig.~\ref{fig:nbc_learn_trav}.
We note that as time goes on and the team experiences more and more of the map, the model becomes increasingly good.
Roads become more well-defined as highly traversable areas and buildings are defined as untraversable areas.
Grassy regions take a value somewhere in between: sometimes drivable, sometimes not.
This suggests that our modeling and analysis system is able to draw useful conclusions about the environment, even one that is much more complex than the simulation.

\section{Conclusion}
In this paper, we developed a system for air-ground collaboration, where robots are able to use information from each other to improve their own localization and planning in the absence of GPS for all robots except the UAV.
Semantics enabled robots diverse in sensing and mobility to represent the world around them and communicate with one another in a common way, even with only intermittent communication.
In addition, semantics proved to be a strong predictor of traversability, enabling UGVs to use the aerial map as a predictor of unvisited regions.
We have also performed extensive ablation experiments in simulation, demonstrating the contribution of each component of our system.
Finally, we have performed real-world experiments across two different challenging environments, urban and rural.
These experiments have comprised nearly 18 kilometers driven autonomously between the 3 UGVs, demonstrating the large-scale operation and robustness of our system.

A key assumption in this work has also been that an aerial orthomap is representative of the environment.
This is true for most outdoor cases but fails to hold in many environments such as those with overhangs or bridges.
The single high-altitude UAV and multiple UGV architecture is not ideal for all applications.
Additionally, dynamic obstacles remain a significant challenge, and there is currently no mechanism for robots to update previously mapped areas, or the UGVs to update the aerial map.
These limitations provide exciting avenues for future work.

\section{Acknowledgement}
The authors would like to thank Dr.~Barbara Dallap Schaer at the Penn Vet New Bolton Center as well as Dr.~Ethan Stump at ARL for their help finding places to test robots in the real world, a task which can be surprisingly difficult.
We would also like to thank Dr.~Priyanka Shah and Alice Kate Li for their invaluable time spent supporting our field experiments, as well as Alex Zhou and Jeremy Wang for building and repairing the many robots used in this work.

We additionally gratefully acknowledge the support of
ARL DCIST CRA W911NF-17-2-0181, 
NSF Grants CCR-2112665, 
ONR grant N00014-20-1-2822, 
ONR grant N00014-20-S-B001, 
NVIDIA, 
and C-BRIC, a Semiconductor Research Corporation Joint University Microelectronics Program program cosponsored by DARPA. 
Ian Miller also acknowledges the support of a NASA Space Technology Research Fellowship.

\bibliographystyle{apalike}
\bibliography{bib.bib}

\end{document}